\documentclass[conference]{IEEEtran}
\IEEEoverridecommandlockouts
\usepackage{cite}
\usepackage{amsmath,amssymb,amsfonts}
\usepackage{algorithmic}
\usepackage{graphicx}
\usepackage{textcomp}
\usepackage{xcolor}

\usepackage[hidelinks]{hyperref}
\usepackage{url}
\usepackage{subcaption}
\usepackage{multirow}
\def\BibTeX{{\rm B\kern-.05em{\sc i\kern-.025em b}\kern-.08em
    T\kern-.1667em\lower.7ex\hbox{E}\kern-.125emX}}
\begin{document}

\title{Visually Impaired Aid using Convolutional Neural Networks, Transfer Learning, and Particle Competition and Cooperation\\
\thanks{This work is supported by the São Paulo Research Foundation - FAPESP (grant \#2016/05669-4)}
}

\author{\IEEEauthorblockN{Fabricio Breve}
\IEEEauthorblockA{\textit{Institute of Geosciences and Exact Sciences} \\
\textit{S\~{a}o Paulo State University (UNESP)}\\
Rio Claro-SP, Brazil \\
fabricio.breve@unesp.br}
\and
\IEEEauthorblockN{Carlos N. Fischer}
\IEEEauthorblockA{\textit{Institute of Geosciences and Exact Sciences} \\
\textit{S\~{a}o Paulo State University (UNESP)}\\
Rio Claro-SP, Brazil \\
carlos.fischer@unesp.br}
}

\maketitle

\begin{abstract}
Navigation and mobility are some of the major problems faced by visually impaired people in their daily lives. Advances in computer vision led to the proposal of some navigation systems. However, most of them require expensive and/or heavy hardware. In this paper we propose the use of convolutional neural networks (CNN), transfer learning, and semi-supervised learning (SSL) to build a framework aimed at the visually impaired aid. It has low computational costs and, therefore, may be implemented on current smartphones, without relying on any additional equipment. The smartphone camera can be used to automatically take pictures of the path ahead. Then, they will be immediately classified, providing almost instantaneous feedback to the user. We also propose a dataset to train the classifiers, including indoor and outdoor situations with different types of light, floor, and obstacles. Many different CNN architectures are evaluated as feature extractors and classifiers, by fine-tuning weights pre-trained on a much larger dataset. The graph-based SSL method, known as particle competition and cooperation, is also used for classification, allowing feedback from the user to be incorporated without retraining the underlying network. 92\% and 80\% classification accuracy is achieved in the proposed dataset in the best supervised and SSL scenarios, respectively.
\end{abstract}



\section{Introduction}
Globally, it is estimated that at least 2.2 billion people have a vision impairment or blindness \cite{WHO2019}. The majority of them are over 50 years old and live in low and middle-income regions \cite{Bourne2017}. Navigation and mobility are among the most critical problems faced by visually impaired persons. There were many advances in computer vision and some proposed navigation systems in the last decade \cite{Lakde2015,Jiang2019,Hoang2017,Tapu2017,Saffoury2016,Poggi2016}. However, many of them require expensive, heavy, and/or not broadly available equipment \cite{Poggi2016,Saffoury2016,Poggi2015,Tapu2017,Hoang2017,Rizzo2017} or require a network connection to a powerful remote server \cite{Lin2017,Jiang2019}. Therefore, the white cane is still the most popular, simplest tool for detecting obstacles due to its low cost and portability \cite{Lakde2015}.

In the recent years, the rise of deep learning methods \cite{Lecun2015,Goodfellow2016,Schmidhuber2015}, especially convolutional neural networks (CNNs), was responsible for many advances in object recognition and image classification \cite{Krizhevsky2012,Cai2016,Howard2017}. CNNs are a class of deep neural networks commonly employed in visual image analysis. However, learning in CNNs usually requires a very large amount of annotated image samples in order to estimate millions of parameters. Annotating images is usually an expensive or time-consuming task, preventing the application of CNNs in problems with limited training data available \cite{Oquab2014}.

Oquab et al. \cite{Oquab2014} showed that image representations learned with CNNs on large-scale annotated datasets can be efficiently transferred to other visual recognition tasks with a limited amount of training data. They reused layers trained on a large dataset to compute mid-level image representation for images in another dataset, leading to significantly improved classification results. This technique, known as transfer learning, was successfully applied in different scenarios \cite{Shin2016,Huynh2016,Gopalakrishnan2017}.

CNNs training phase commonly has very high computational costs. However, once trained, CNNs are relatively fast to make inferences. Most current smartphones SoCs (System-on-a-Chip) are able to make inferences on a single image using CNN models, like VGG19 \cite{Simonyan2015} and InceptionV3 \cite{Szegedy2016}, in the range of milliseconds \cite{Ignatov2019}.

Considering the current scenario, we glimpse a system to assist visually impaired people that executes on a single smartphone, without extra accessories or connection requirements. It could work added to the white cane, taking pictures of the path a person is walking through, and providing audio and/or vibration feedback regarding potential obstacles, even before they are in the reach of the white cane.

In this paper, we propose a dataset of images taken of the path using a smartphone camera in a first-person perspective. Though not large, the dataset covers indoor and outdoor situations, different types of floor, sidewalks with art on the floor, dry and wet floor, different amounts of light, daylight and artificial light, and different types of obstacles.

We also propose a CNN classifier to label these images in two classes: ``clear path'' and ``non-clear path''. It is based on transfer learning, by using CNN models with weights pre-trained on the ImageNet dataset \cite{ILSVRC15}, which has millions of images and hundreds of classes. The weights of specific blocks of layers can then be fine-tuned to the proposed dataset. It is expected that these networks can learn and extract useful features in spite of the dataset challenges. We have tested $17$ different CNN architectures to find out which ones would perform better on this task, and Xception \cite{Chollet2017}, VGG16 \cite{Simonyan2015}, VGG19 \cite{Simonyan2015}, InceptionV3 \cite{Szegedy2016}, and MobileNet \cite{Howard2017} achieved the best results.


Moreover, we glimpse at the possibility of incorporating knowledge from user feedback, who could point out wrong inferences and/or confirm the right ones. In this case, since it is not feasible to re-train or fine-tune the network on the smartphone, an alternative approach is proposed using the graph-based semi-supervised learning (SSL) method known as particle competition and cooperation (PCC) \cite{Breve2012}. In this scenario, VGG16 and VGG19 models \cite{Simonyan2015}, without the classification layer, work as feature extractors. They were chosen among the others because they are the only architectures that achieved good results without fine-tuning. These CNN models were designed to attend a competition of large scale visual recognition on the ImageNet dataset in 2014. Principal Component Analysis (PCA) \cite{Jolliffe2002} is used to reduce the output of the last CNN convolutional layer. A few of the principal components are then used to build a graph, in which each image is a node and edges connect each of them to their closest neighbors. The graph is fed to PCC with a few annotated images and it performs the classification of the remaining.

The remaining of this paper is organized as follows: Section \ref{sec:relatedwork} shows some related work on visually impaired aid, transfer learning, and the PCC model. Section \ref{sec:dataset} presents the proposed dataset. Section \ref{sec:baselineCNN} introduces the baseline CNN, a first attempt to classify the proposed dataset, used for comparison with other proposed approaches. Section \ref{sec:TransferLearning} presents our proposed architecture using transfer learning, with computer simulations and their corresponding results. Section \ref{sec:PCC} presents our experiments with SSL, using CNNs as feature extractors and PCC for classification. Finally, in Section \ref{sec:Conclusions} we draw some conclusions.

\section{Related Work}
\label{sec:relatedwork}

In this section, we briefly describe some related work, focused on visually impaired aid, transfer learning, and the particle competition and cooperation model.

\subsection{Visually Impaired Aid}

Many efforts have been made to incorporate computer vision technologies to assist visually impaired people. Kumar and Meher \cite{Kumar2015} presented an object detection system, using a combination of CNN and RNN (Recurrent Neural Network). It recognizes common indoor objects and their colors, providing audio feedback to the user. Saffoury et al. \cite{Saffoury2016} propose a system with a laser pointer and a smartphone. They used laser triangulation to build a collision avoidance algorithm. They also provide audio feedback to the user.

Poggi and Mattoccia \cite{Poggi2016} built a wearable device, with glasses with a custom RGBD sensor and FPGA onboard processing, a glove with micro-motors for tactile feedback, a pocket battery, a bone-conductive headset, and a smartphone. They used deep-learning techniques to semantically categorize detected obstacles. Earlier, Poggi et al. \cite{Poggi2015} have presented a similar system for crosswalk recognition.

Tapu et al. \cite{Tapu2013} presented a real-time obstacle detection and classification system, using video captured from a smartphone camera. They built a framework including tracking, motion estimation, and clustering techniques. Four years later, Tapu et al. \cite{Tapu2017} introduced a more complex framework, based on CNNs to detect, track and recognize objects in an outdoor environment. However, the system requires a laptop computer, carried on a backpack, to be used as a processing unit.

Hoang et al. \cite{Hoang2017} presented a system that uses a Kinect to capture the environment. It detects obstacles and provides audio feedback. It also requires a laptop computer to be carried on a backpack. Rizzo et al. \cite{Rizzo2017} propose a fusion framework to join merge signals from a stereo camera and an infrared sensor. The obstacle detection is performed using a CNN.

Lin et al. \cite{Lin2017} propose a smartphone-based guiding system. They also use CNNs for object recognition. However, they rely on a desktop server with a GPU with Compute Unified Device Architecture (CUDA) to perform the heavier object recognition task. The system has an offline mode but it only provides face and stairs recognition. Jiang et al. \cite{Jiang2019} propose a wearable system based on stereo vision. They used a CNN on the cloud for image recognition.

Islam and Sadi \cite{Islam2018} used a CNN for path hole detection. Though they achieved very good results, it is worth noticing that they used a dataset for the `path hole' images and another one for `non-path hole' images. The later consists of road pictures taken with a wider angle. Therefore, it is likely that the network also learned these style differences that are not related to the problem, making its job easier.

\subsection{Transfer Learning}

CNNs usually require a large amount of training data to effectively learn. However, when limited training data is available, it is possible to efficiently transfer representations learned by a CNN architecture on other visual recognition tasks, in which large-scale datasets are available.

Oquab et al. \cite{Oquab2014} reused layers trained on the ImageNet dataset to compute mid-level image representation for images and classify objects in the Pascal VOC dataset \cite{Everingham2010}, outperforming the previous state-of-the-art results. Gopalakrishnan et al. \cite{Gopalakrishnan2017} employed CNNs trained on ImageNet to detect cracks in surface pavement images. Shin et al. \cite{Shin2016} successfully applied transfer learning from the natural images of the ImageNet dataset to computed tomography images of the medical domain.

Saleh et al. \cite{Saleh2017} used a CNN based on VGG16 to detect navigational path, through segmentation of images on a pixel-wise level. They have added some convolutional and de-convolutional layers which are specific for their segmentation task. But they also fine-tuned pre-trained weights on the layers based on VGG16.

Monteiro et al. \cite{Monteiro2017} used a dataset of videos taken from the point of view of guide-dog to train a CNN to recognize activities taking place around the user camera. They made simulations with both fully trained and fine-tuned AlexNet \cite{Krizhevsky2012} and GoogLeNet \cite{Szegedy2015} networks pre-trained on the ILSVRC 2012 ImageNet dataset \cite{ILSVRC15}.

\subsection{Particle Competition and Cooperation}

Particle competition and cooperation (PCC) \cite{Breve2012} is a nature-inspired graph-based semi-supervised learning (SSL) method. In this model, teams of particles walk through a network, cooperating among themselves and competing against particles from other teams, trying to possess as many nodes as possible. The teams are the classes in the machine learning context. Therefore, particles representing the same class walk cooperatively to spread their label. At the same time, particles from different classes compete to define the classes' boundaries.

The network is represented as a graph and is generated from the input data. Each element becomes a graph node, and the edges are created between each node and its $k$-nearest neighbors, given by some distance measured among them, usually the Euclidean distance. For each node that corresponds to a pre-labeled element, a particle is generated, whose initial position is the same node, known as the ``home node'' of the particle. As the particles change position, the distance between their current node and their home node is registered. Particles generated by elements of the same class act as a team.

Each graph node has a vector where each element represents the domination level of a team over that node. The sum of the vector is always constant. As the system executes, particles walk through the graph and raise the dominance level of their team over the node, at the same time that they lower other team domination levels, always keeping the constant sum. Moreover, each particle also has a strength level, which raises when it visits a node dominated by its team and lowers when it visits a node dominated by another team. This strength is important because the change a particle causes in a node is proportional to the strength it has at the time. This mechanism ensures that a particle is stronger when it is in its neighborhood, protecting it, and it is weaker when it is trying to invade another team territory.

Particles choose the next node to be visited based on one of two rules. At each iteration, they randomly choose one of the rules with pre-defined probabilities. The two rules are described as follows:
\begin{itemize}
  \item \textbf{Random rule}: the particle randomly chooses, with equal probabilities, any neighbor node to visit.
  \item \textbf{Greedy rule}: the particle randomly chooses any neighbor node to visit but with probabilities proportional to the dominance level its team has on each neighbor and inversely proportional to the distance of the neighbor to their home node.
\end{itemize}

Therefore, the greedy rule is useful to keep the particles in their own territory, i.e., defensive behavior. On the other hand, using the random rule the particles are more likely to go to non-dominated and distant nodes, assuming an explorative behavior.

Notice that a particle only stays on the chosen node if it is able to dominate that node, otherwise, it is expelled and goes back to the previous node until the next iteration. This rule is used to avoid that a particle leaves its territory behind and loses all its strength. It also favors smooth borders on the territories, since a particle cannot dominate a given node before dominating the nodes on its path. At the end of the iterations, each node is labeled after the team that has dominated it.

PCC was already extended to handle fuzzy community structure \cite{Breve2013SoftComputing}, to be more robust to label noise \cite{Breve2015Neurocomputing}, to handle data streams and concept drift \cite{Breve2012IJCNN}, to perform active learning \cite{Breve2013IJCNN} and image segmentation \cite{Breve2015IJCNN}, among others. It has been applied to data from different domains, including software engineering, bioinformatics, and medical diagnostics.

\section{Proposed Dataset}
\label{sec:dataset}

The proposed dataset\footnote{The dataset is available at: \url{https://github.com/fbreve/via-dataset}} consists of $342$ images divided into two classes: $175$ of them are ``clear-path'' and $167$ are ``non-clear'' path. They were taken using a smartphone camera and resized to $750 \times 1000$ pixels. The smartphone was placed in the user chest height and inclined approximately 30º to 60º from the ground, so it could capture a few meters of the path ahead, and beyond the reach of a regular white cane.

\begin{figure}
\centering
\begin{subfigure}[b]{\columnwidth}
\centering
\includegraphics[width=.24\columnwidth]{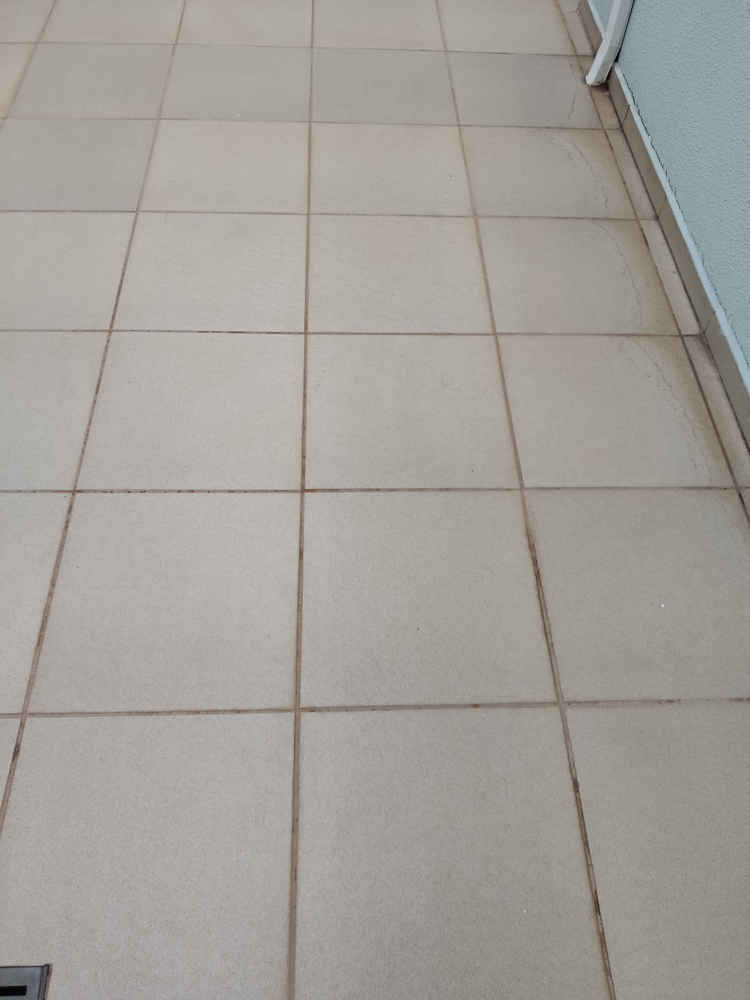}
\includegraphics[width=.24\columnwidth]{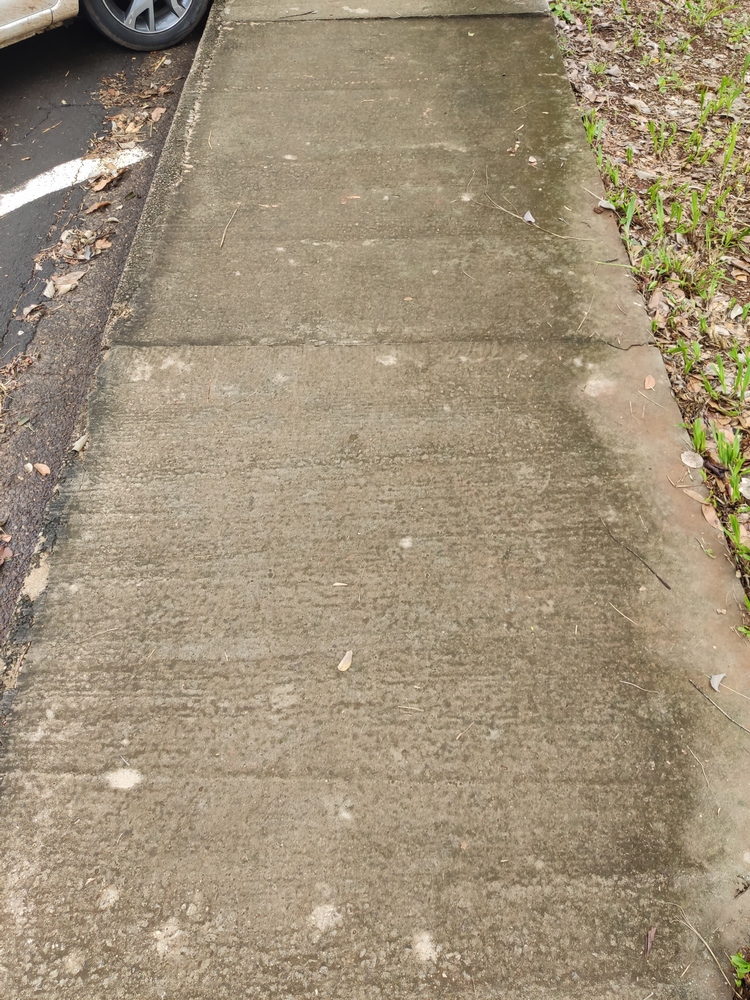}
\includegraphics[width=.24\columnwidth]{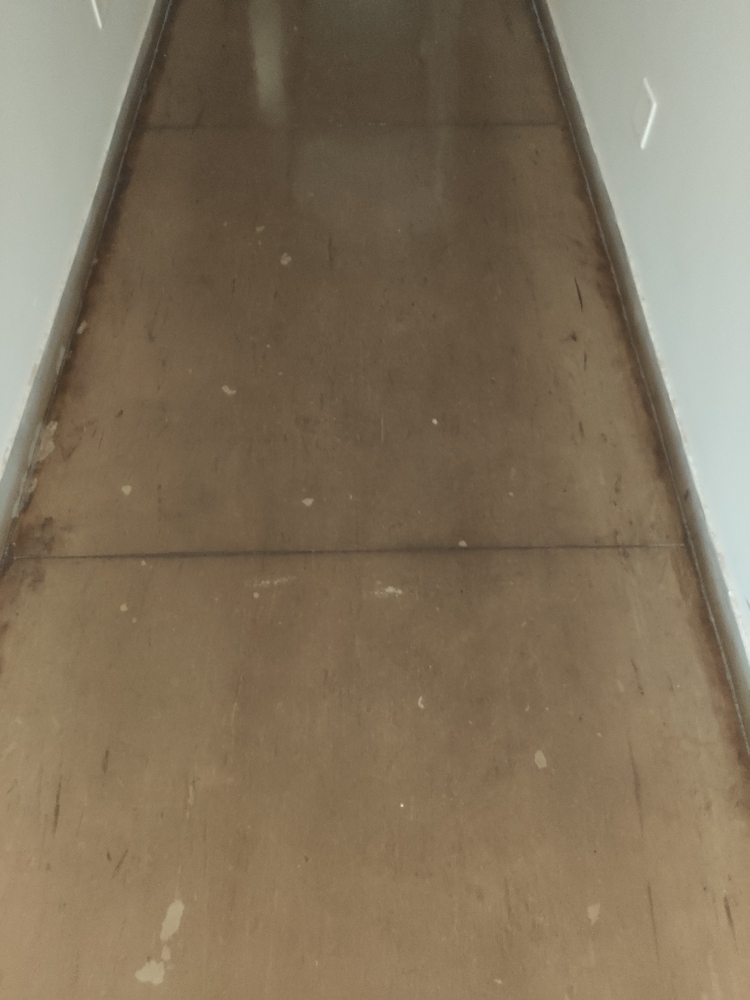}
\includegraphics[width=.24\columnwidth]{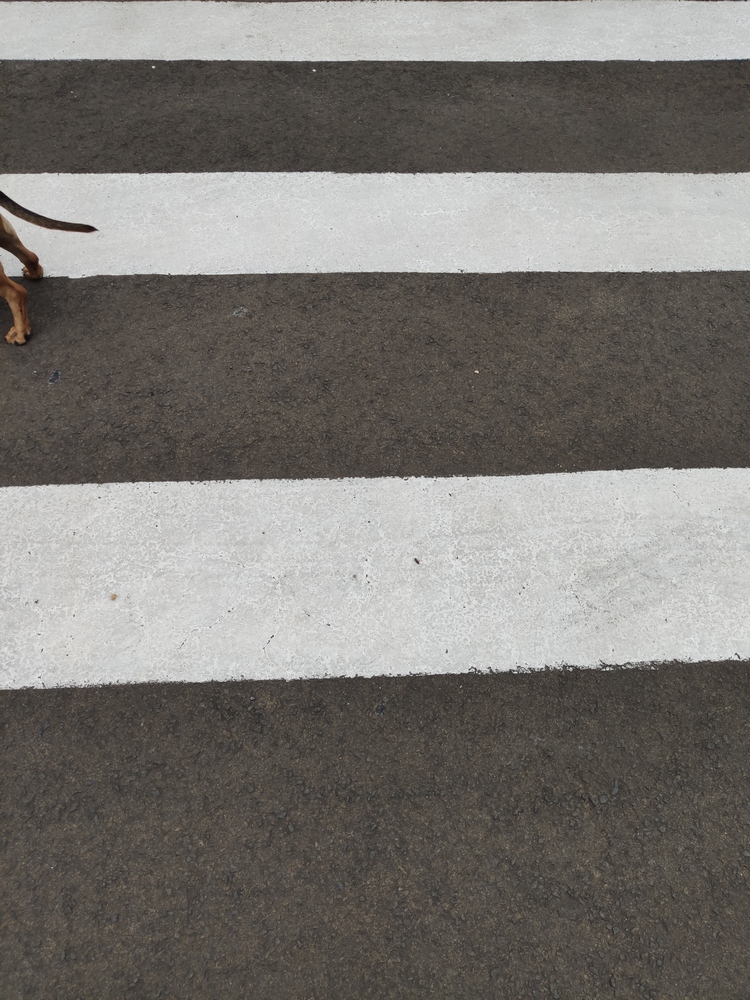} \\
\vspace{0.015\columnwidth}
\includegraphics[width=.24\columnwidth]{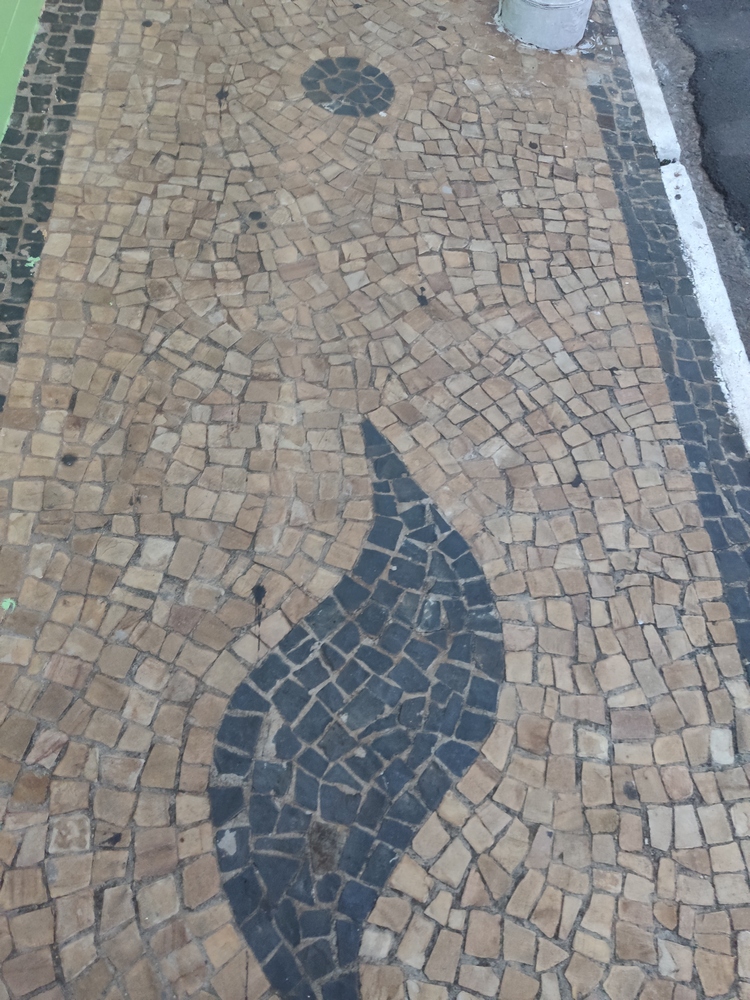}
\includegraphics[width=.24\columnwidth]{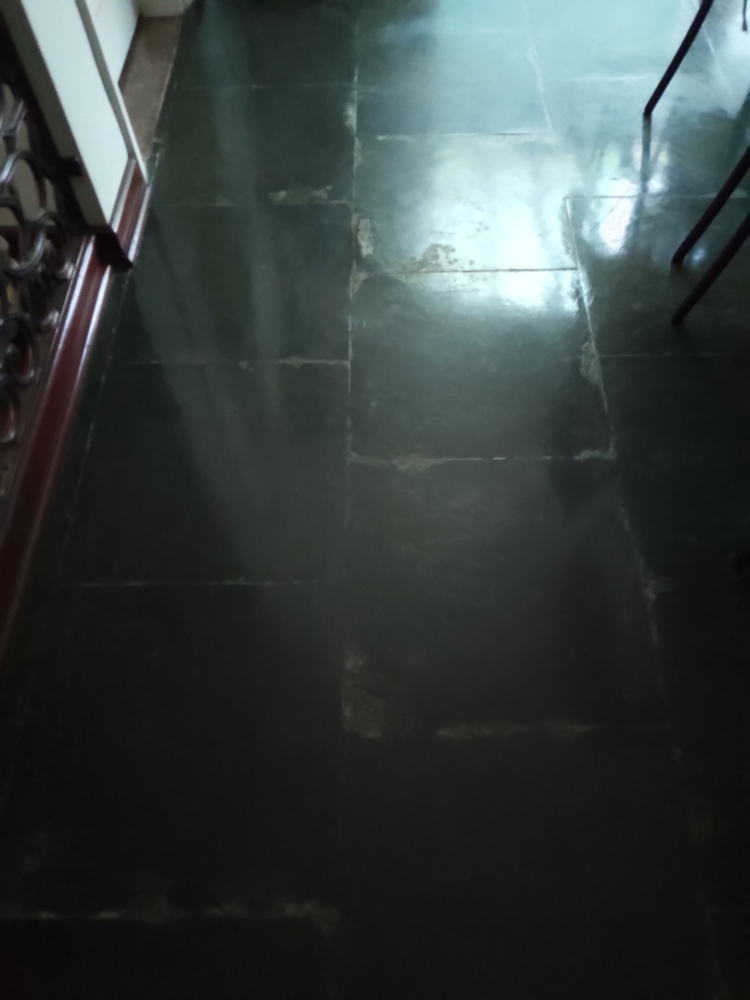}
\includegraphics[width=.24\columnwidth]{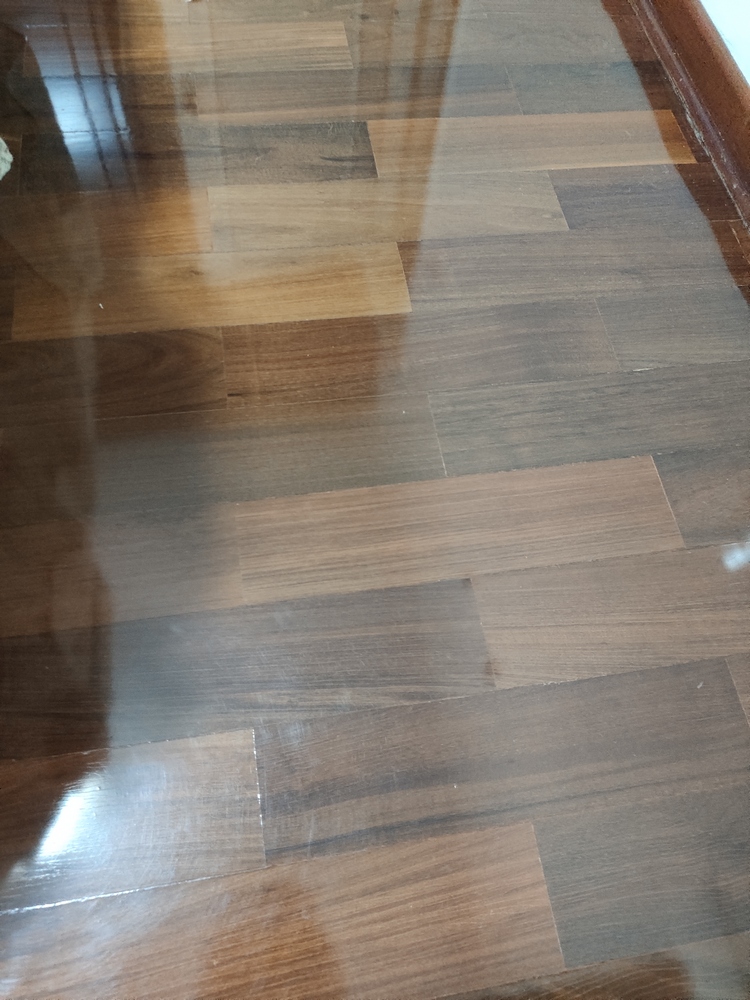}
\includegraphics[width=.24\columnwidth]{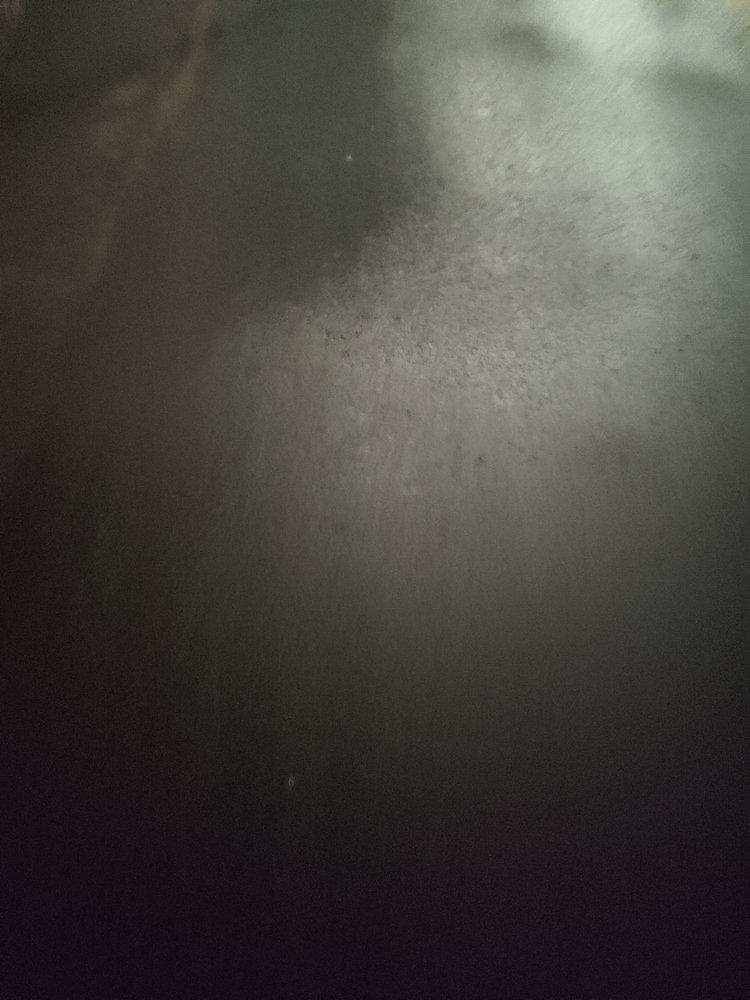}
\caption{``clear path''}
\end{subfigure}

\vspace{0.03\columnwidth}

\begin{subfigure}[b]{\columnwidth}
\centering
\includegraphics[width=.24\columnwidth]{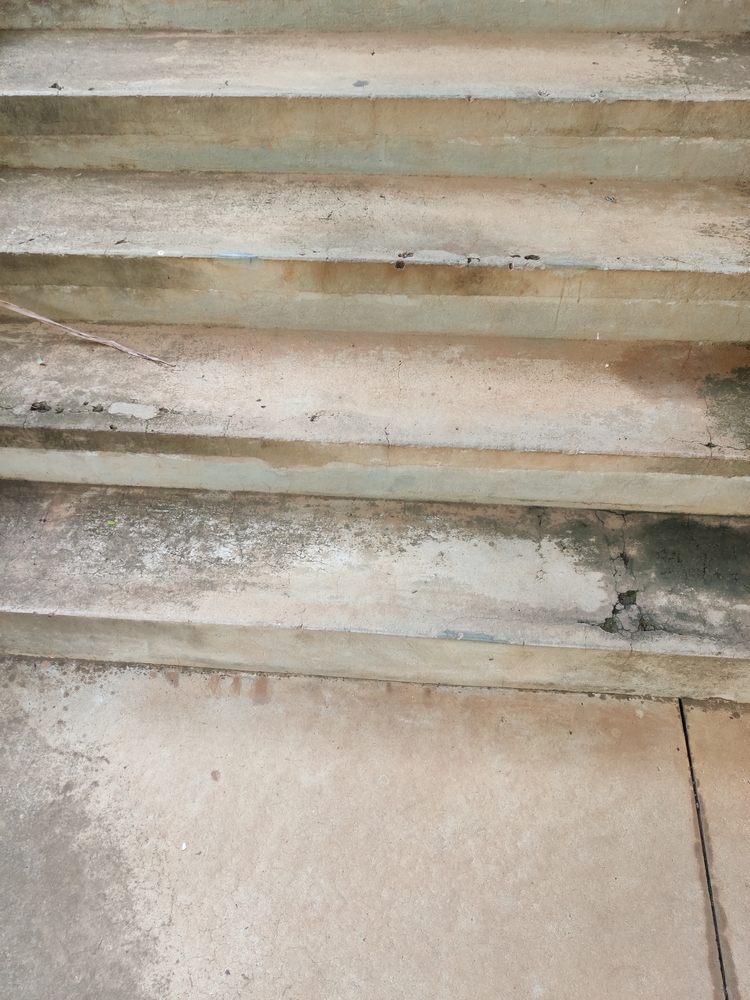}
\includegraphics[width=.24\columnwidth]{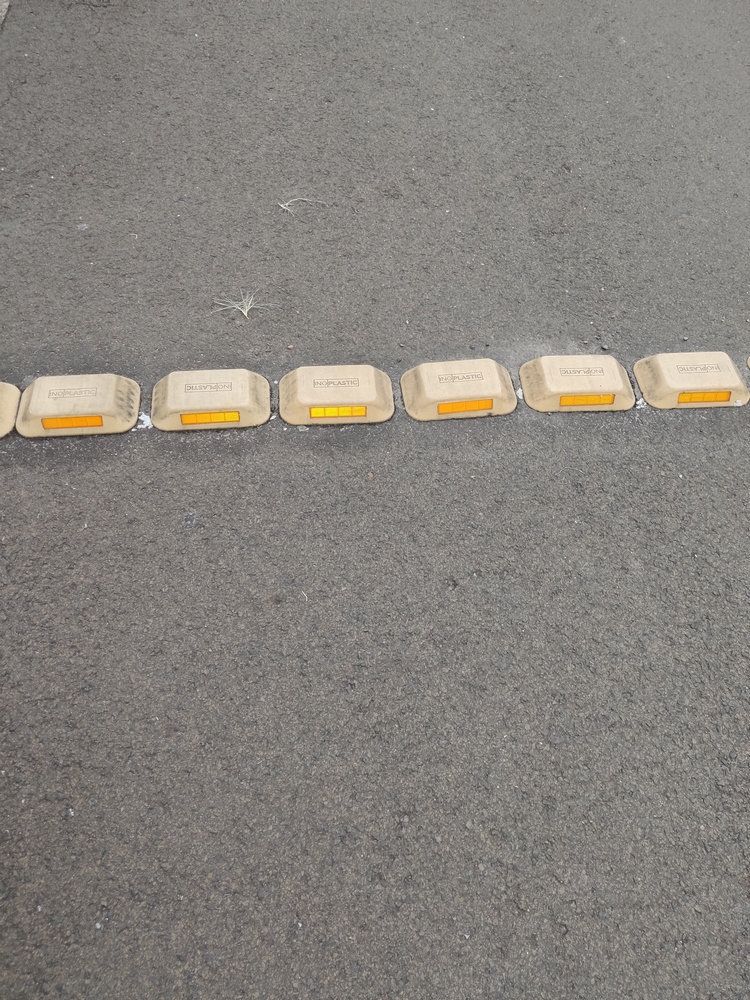}
\includegraphics[width=.24\columnwidth]{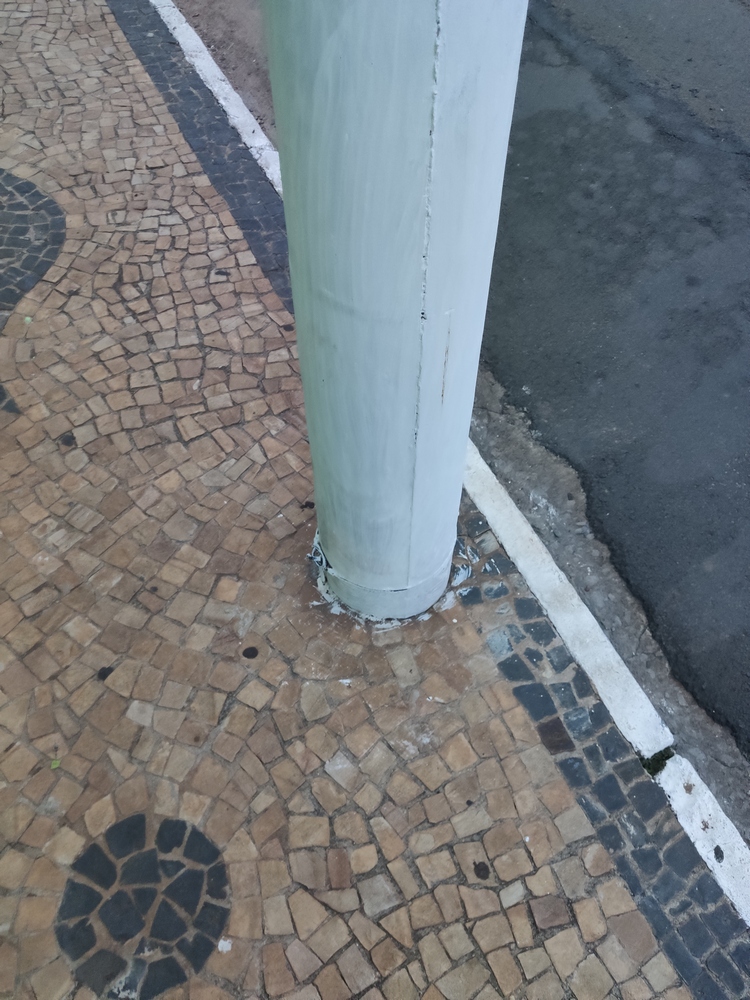}
\includegraphics[width=.24\columnwidth]{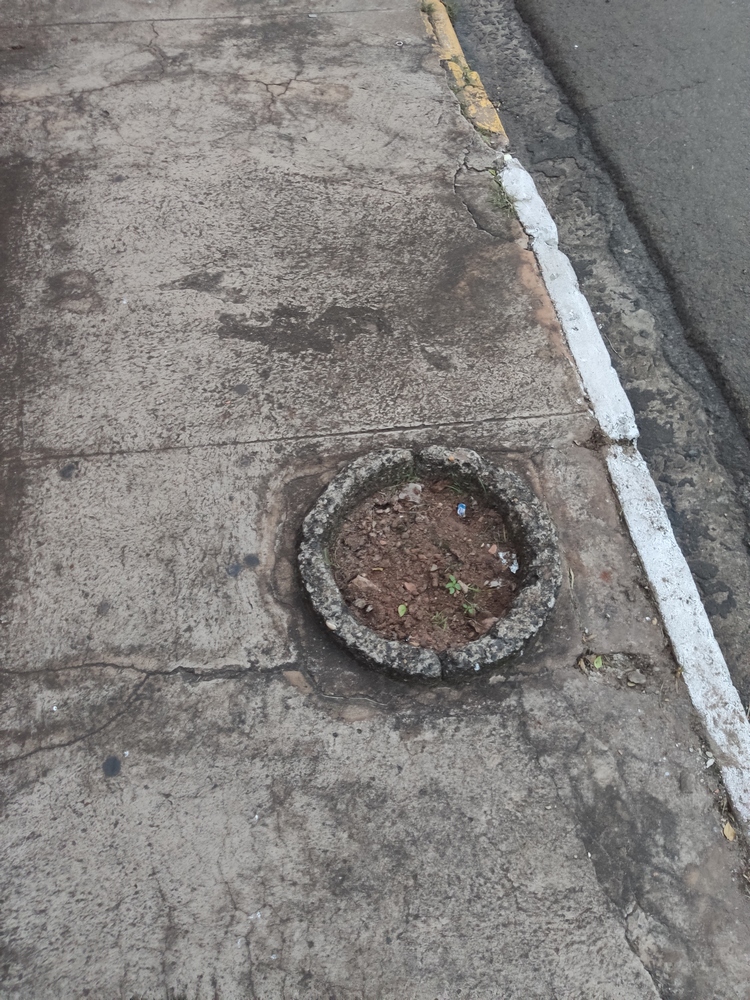} \\
\vspace{0.015\columnwidth}
\includegraphics[width=.24\columnwidth]{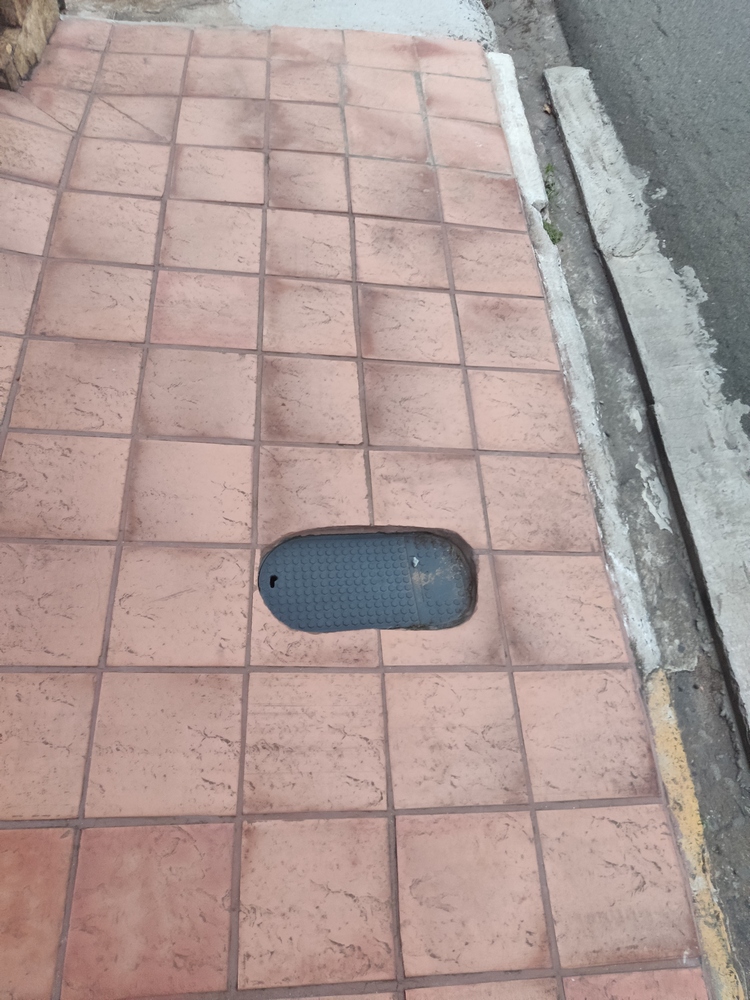}
\includegraphics[width=.24\columnwidth]{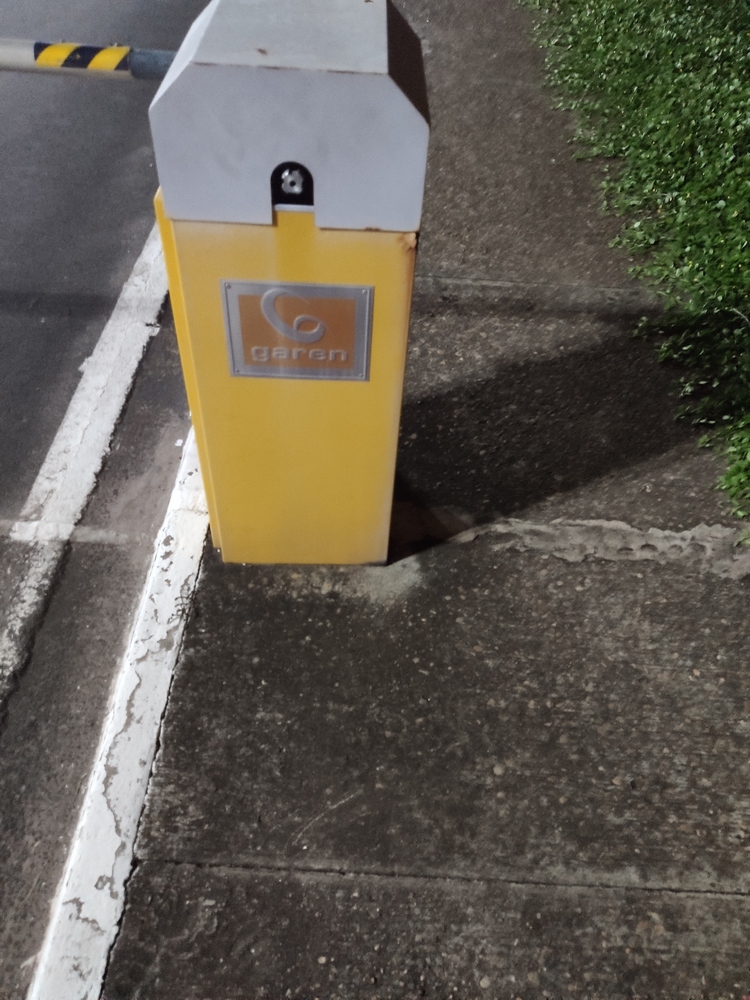}
\includegraphics[width=.24\columnwidth]{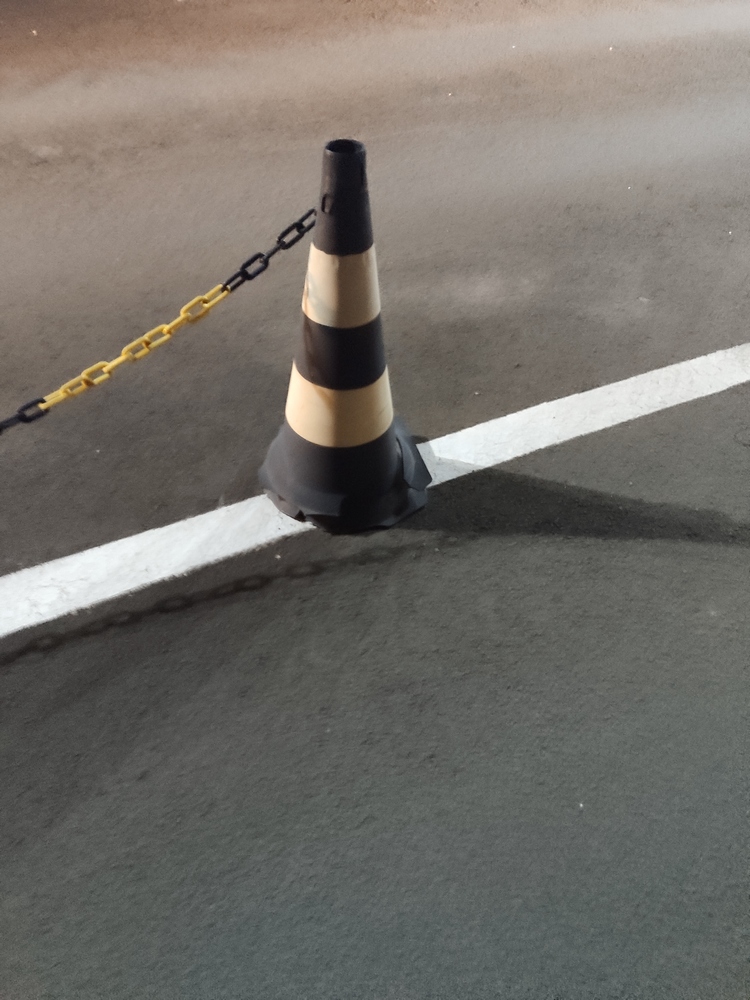}
\includegraphics[width=.24\columnwidth]{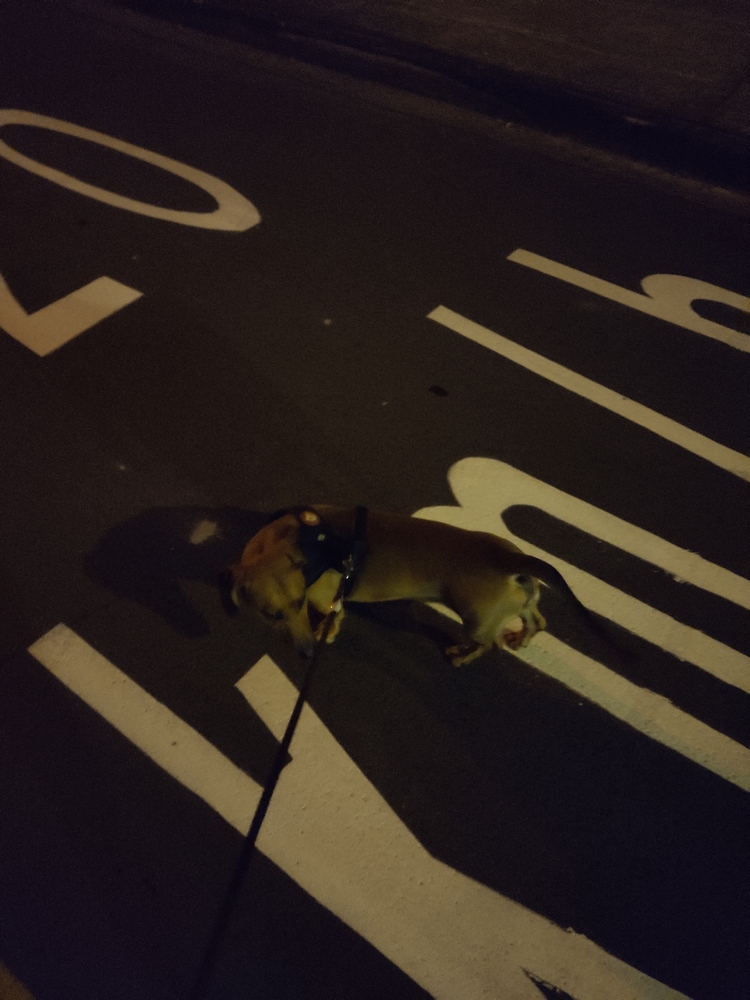}
\caption{``non-clear path''}
\end{subfigure}
\caption{Some images extracted from the proposed dataset: (a) ``clear path'' class; and (b) ``non-clear path'' class.}
\label{fig:dataset}
\end{figure}

Though not large, the dataset covers indoor and outdoor situations, with different types of floor, including both dry and wet floor; different amounts of light, including both daylight and artificial light; and different types of obstacles, like stairs, trees, holes, animals, traffic cones, among others. Fig.~\ref{fig:dataset} shows some examples of images from the proposed dataset.

\section{Baseline CNN}
\label{sec:baselineCNN}

\begin{figure}
\centering
\includegraphics[width=8.5cm]{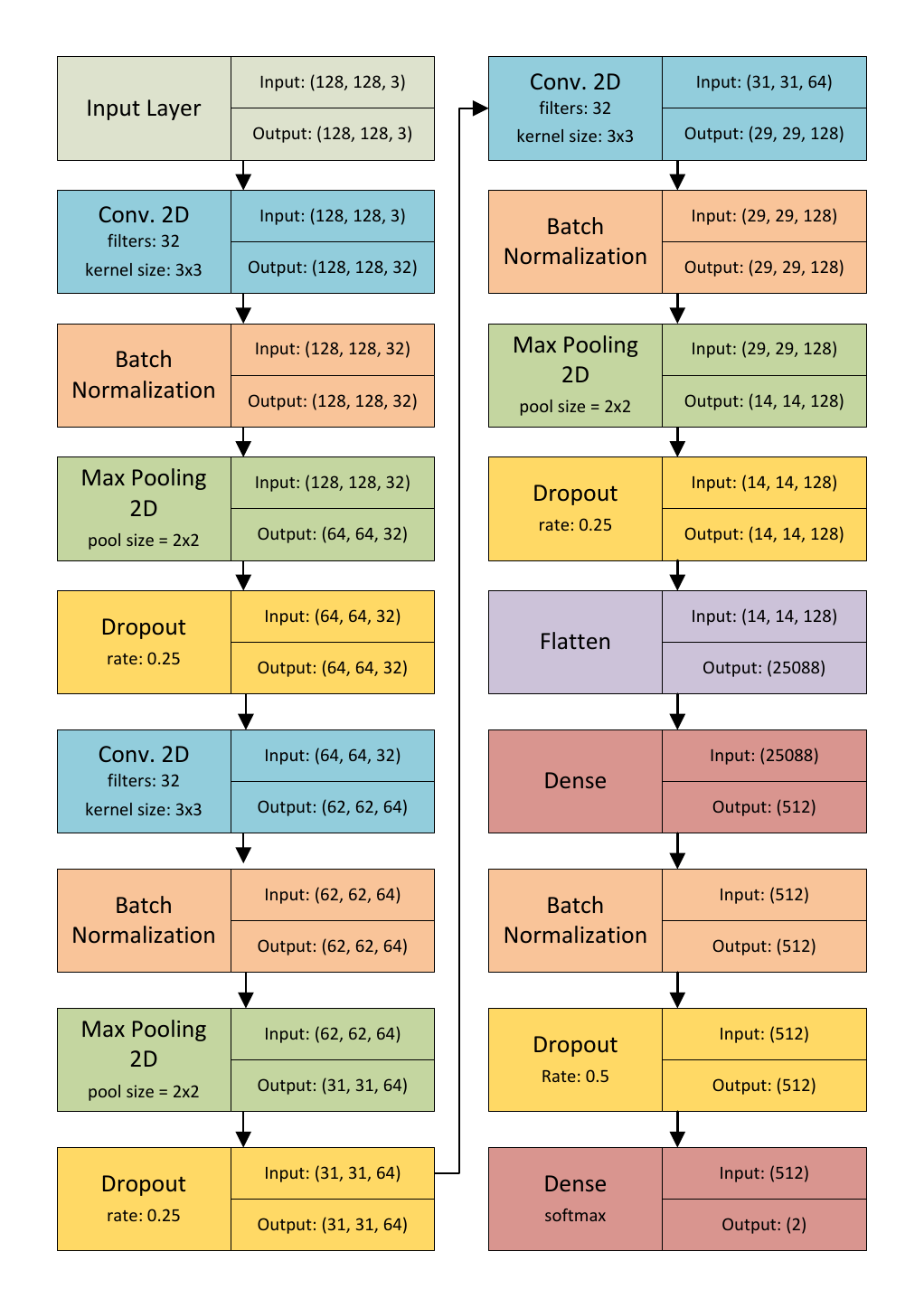}
\caption{Baseline CNN diagram.}
\label{fig:baselineCNNdiagram}
\end{figure}

For the first attempt to classify the proposed dataset, we build a CNN with three convolutional layers. Each layer is followed by a batch normalization layer, a max-pooling layer, and a dropout layer. After the third layer, there is a dense intermediate layer, also followed by batch normalization and dropout and, finally, a dense classification layer with two outputs: ``clear'' and ``non-clear''. All activation functions are \emph{ReLU (Rectified Linear Unit)}, except for the classification layer, which uses \emph{softmax}. The input images are resized to $128 \times 128$ pixels, for faster processing. This CNN is used as a baseline for later computer simulations; its full diagram is displayed in Fig.~\ref{fig:baselineCNNdiagram}.

The network is implemented in Python, using the TensorFlow framework. It is trained with three different optimizers: Adam \cite{Kingma2014,Reddi2018}, RMSprop \cite{Tieleman2012}, and SGD (Stochastic Gradient Descent). The option \emph{earlystop} is used in all scenarios, with the \emph{patience} parameter set to $10$. \emph{learning\_rate\_reduction} is also used, with reduction on plateau and the following parameters: \emph{patience} = $2$, \emph{factor} = $0.5$, and \emph{min\_lr} = $0.00001$.

In some simulations, we used real-time data augmentation, i.e., the images are presented more than once per-epoch and with slightly random transformations to help the network to learn and to generalize. This is specially useful on small datasets. The following parameters were used: \emph{rotation\_range} = $15$, \emph{shear\_range} = $0.1$, \emph{zoom\_range} = $0.2$, \emph{horizontal\_flip} = True, \emph{width\_shift\_range} = $0.1$, \emph{and height\_shift\_range} = 0.1.

Table \ref{tab:CNNBaselineResults} shows the classification accuracy obtained with the proposed network, using the different optimizers and with and without data augmentation. All results were obtained using K-Fold Cross Validation with $k=10$; the process is repeated $10$ times, so each value on Table \ref{tab:CNNBaselineResults} is the average of $100$ executions.

\begin{table}
\caption{Classification accuracy obtained with the Baseline CNN network, with and without data augmentation, and using different optimizers. The best result is highlighted in bold.}
\centering
\begin{tabular}{ccrl}
\hline
{\bf Data Augmentation} & {\bf Optimizer} & \multicolumn{2}{c}{\bf Accuracy} \\
\hline
    no & Adam       & $67.97\%$     & $\pm 7.33\%$ \\
    no & RMSProp    & $70.35\%$     & $\pm 8.67\%$  \\
    no & SGD        & $60.53\%$     & $\pm 8.64\%$ \\
    yes & Adam      & $73.51\%$     & $\pm 7.98\%$ \\
    yes & RMSProp   & $\bf{76.40\%}$ & $\bf{\pm 7.14\%}$ \\
    yes & SGD       & $72.19\%$     & $\pm 7.57\%$    \\
\hline
\end{tabular}
\label{tab:CNNBaselineResults}
\end{table}

The best accuracy (76.40\%) is obtained with data augmentation and the RMSProp optimizer. Based on these results, we build another set of simulations to explore transfer learning.

\section{CNN Architectures and Transfer Learning}
\label{sec:TransferLearning}

\begin{figure}
\centering
\begin{subfigure}[c]{0.49\columnwidth}
    \centering
    \includegraphics[width=\linewidth]{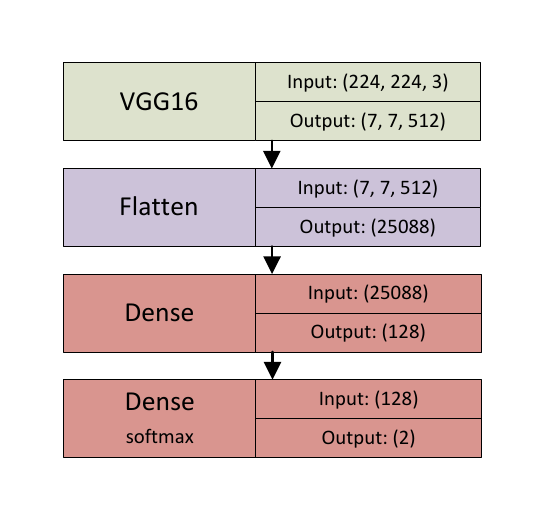}
    \caption{No Pooling}
\end{subfigure}
\begin{subfigure}[c]{0.49\columnwidth}
    \centering
    \includegraphics[width=\linewidth]{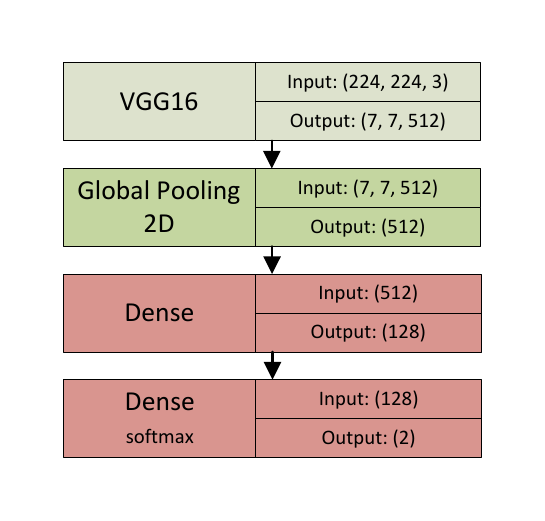}
    \caption{2D Global Pooling}
\end{subfigure}
\caption{Proposed Transfer Learning networks using VGG16 - Model A: (a) No Pooling; (b) 2D Global Pooling.}
\label{fig:TLmodelAdiagram}
\end{figure}

In this section, we explore transfer learning on the proposed dataset. We propose a simple model where the output of the last convolutional layer of the original CNN is flattened and passed to a small intermediate dense layer, followed by a classification layer. Alternatively, a global pooling layer may be used instead of the flatten layer. In the VGG16 case, the output is a 3D matrix $7 \times 7 \times 512$, which becomes a single $512$ vector when using global pooling. These models are presented in Fig.~\ref{fig:TLmodelAdiagram}, taking VGG16 \cite{Simonyan2015} as example of CNN architecture. In our experiments, we used no pooling and global average pooling.

In the first set of transfer learning computer simulations, we explore $17$ different CNN architectures. In all scenarios, the original networks were pre-trained in the ImageNet dataset \cite{ILSVRC15}, a very large dataset with millions of images and hundreds of classes. All the pre-trained weights of the networks are available in Tensorflow. The classification layer of the original CNN is removed, and the output of its last convolutional layer is fed to our model. We tested four different scenarios: in the first two, the weights of the original CNN layers are frozen and only the layers we add are trained. Therefore, we might say that the original CNNs work as feature extractors. In the first scenario, we use no pooling and in the second one, we use global average pooling. In the third and fourth scenarios, all the weights are fine-tuned to the proposed dataset. Again, no pooling is used in the third one and global average pooling is used in the forth.

Table \ref{tab:TLallCNN} shows the classification accuracy achieved using transfer learning with the $17$ different CNN architecture. In all cases, RMSProp is the chosen optimizer, as it achieved the best results in the previous simulations. No data augmentation is used and the other parameters are the same used with the baseline CNN. All dataset images are resized to the default input size of each method (either $224 \times 224$ or $299 \times 299$ pixels in these CNNs). All the results were obtained using K-Fold Cross Validation with $k=10$, so each value on the table is the average of $10$ executions.

\begin{table*}
\caption{Classification accuracy obtained using transfer learning with $17$ different CNN architectures. The best results in each column are in bold.}
\centering
\begin{tabular}{|c|rl|rl|rl|rl|}
\hline
\multirow{2}*{\bf Architecture} & \multicolumn{4}{|c|}{\bf Frozen Weights} & \multicolumn{4}{c|}{\bf Fine-Tunable Weights} \\
\cline{2-9}
 & \multicolumn{2}{c|}{\bf No Pooling} & \multicolumn{2}{c|}{\bf Average Pooling} & \multicolumn{2}{c|}{\bf No Pooling} & \multicolumn{2}{c|}{\bf Average Pooling} \\
\hline
Xception \cite{Chollet2017}             & $49.43\%$ & $ \pm 6.82\%$             & $51.18\%$ & $\pm 8.61\%$  &  $\bf{87.70\%}$ & $\bf{\pm 2.62\%}$ & $\bf{92.11\%}$ & $\bf{\pm 5.01\%}$\\
VGG16 \cite{Simonyan2015}               & $\bf{83.39\%}$ & $\bf{\pm 13.35\%}$   & $\bf{76.88\%}$ & $\bf{\pm 7.32\%}$  &  $85.76\%$ & $ \pm 11.97\%$        & $85.16\%$ & $\pm 12.96\%$ \\
VGG19 \cite{Simonyan2015}               & $81.36\%$ & $ \pm 11.11\%$            & $73.97\%$ & $\pm 5.53\%$  &  $83.40\%$ & $ \pm 11.38\%$         & $85.18\%$ & $ \pm 13.29\%$ \\
ResNet50 \cite{He2016}                  & $51.17\%$ & $ \pm 8.82\%$             & $51.18\%$ & $\pm 8.61\%$  &  $49.98\%$ & $ \pm 9.31\%$          & $51.77\%$ & $ \pm 8.77\%$\\
ResNet101 \cite{He2016}                 & $51.18\%$ & $ \pm 8.61\%$             & $50.87\%$ & $\pm 8.08\%$  &  $66.12\%$ & $ \pm 19.75\%$         & $51.42\%$ & $ \pm 8.83\%$ \\
ResNet152 \cite{He2016}                 & $48.05\%$ & $ \pm 10.97\%$            & $51.17\%$ & $\pm 7.88\%$  &  $54.90\%$ & $ \pm 12.51\%$         & $47.37\%$ & $ \pm 4.85\%$\\
ResNet50V2 \cite{He2016ResNetV2}        & $51.19\%$ & $ \pm 11.08\%$            & $51.19\%$ & $\pm 11.08\%$ &  $51.48\%$ & $ \pm 8.72\%$          & $69.71\%$ & $ \pm 22.50\%$ \\
ResNet101V2 \cite{He2016ResNetV2}       & $51.18\%$ & $ \pm 8.61\%$             & $51.18\%$ & $\pm 8.12\%$  &  $63.49\%$ & $ \pm 9.02\%$          & $51.46\%$ & $ \pm 9.02\%$ \\
ResNet152V2 \cite{He2016ResNetV2}       & $51.18\%$ & $ \pm 8.12\%$             & $53.25\%$ & $\pm 11.63\%$ &  $54.40\%$ & $ \pm 7.53\%$          & $54.69\%$ & $ \pm 6.24\%$ \\
InceptionV3 \cite{Szegedy2016}          & $51.18\%$ & $ \pm 8.61\%$             & $51.18\%$ & $\pm 8.61\%$  &  $79.94\%$ & $ \pm 16.02\%$         & $88.90\%$ & $ \pm 3.38\%$ \\
InceptionResNetV2 \cite{Szegedy2017}    & $51.18\%$ & $ \pm 8.12\%$             & $51.19\%$ & $\pm 11.08\%$ &  $75.53\%$ & $ \pm 11.90\%$         & $78.37\%$ & $ \pm 5.08\%$\\
MobileNet \cite{Howard2017}             & $51.48\%$ & $ \pm 8.72\%$             & $51.18\%$ & $\pm 8.61\%$  &  $81.43\%$ & $ \pm 16.66\%$         & $90.08\%$ & $ \pm 5.02\%$ \\
DenseNet121 \cite{Huang2017}            & $51.18\%$ & $ \pm 8.61\%$             & $51.18\%$ & $\pm 8.61\%$  &  $54.71\%$ & $ \pm 9.75\%$          & $45.03\%$ & $ \pm 8.62\%$ \\
DenseNet169 \cite{Huang2017}            & $51.45\%$ & $ \pm 8.04\%$             & $48.24\%$ & $\pm 8.73\%$  &  $64.63\%$ & $ \pm 12.02\%$         & $75.50\%$ & $ \pm 15.18\%$ \\
DenseNet201 \cite{Huang2017}            & $48.34\%$ & $ \pm 10.90\%$            & $51.46\%$ & $\pm 9.11\%$  &  $61.80\%$ & $ \pm 17.82\%$         & $59.34\%$ & $ \pm 8.70\%$ \\
NASNetMobile \cite{Zoph2018}            & $51.78\%$ & $ \pm 9.43\%$             & $49.13\%$ & $\pm 12.40\%$ &  $50.29\%$ & $ \pm 8.46\%$          & $49.13\%$ & $ \pm 8.80\%$\\
MobileNetV2 \cite{Sandler2018}          & $50.90\%$ & $ \pm 8.65\%$             & $51.19\%$ & $\pm 11.08\%$ &  $50.90\%$ & $ \pm 8.65\%$          & $51.18\%$ & $ \pm 8.61\%$\\
\hline
\end{tabular}
\label{tab:TLallCNN}
\end{table*}

From Table \ref{tab:TLallCNN} we observe that, with frozen weights, only VGG16 and VGG19 architectures were able to outperform the baseline CNN. The other architectures had bad results, suggesting that they are not good extractors for this task without some kind of fine-tuning. When all the weights are fine-tuned, Xception, InceptionV3, InceptionResNetV2, and MobileNet were able to learn and outperform the baseline CNN. VGG16 and VGG19 also had improved results with fine-tuning. Global average pooling led to better results when using fine-tuning, while no pooling is better when the weights are frozen.

Sometimes, freezing the first layers and fine-tuning the last ones may improve the classification accuracy. Usually, the first convolutional layers extract low-level features, common in many kinds of images. Therefore, they may not require adjustments. On the other hand, the last convolutional layers extract high-level problem-specific features and are less likely to be useful without fine-tuning. Thus, it is worth investigating how some models behave when the initial layers are frozen and the remaining are fine-tuned.

VGG16 has $5$ blocks of convolutional layers. Table \ref{tab:TLVGG16} shows the classification accuracy with different amounts of frozen blocks of layers, starting with all blocks being fine-tuned and then increasingly freezing each block from the network input to the output. All the results were obtained using K-Fold Cross Validation with $k=10$; the whole process is repeated $10$ times, so each value on the table is the average of $100$ executions. The best results were achieved with 3 frozen blocks with no pooling and 2 frozen blocks with global average pooling.

\begin{table}
\caption{Classification accuracy obtained using transfer learning with VGG16 and different amounts of frozen blocks of layers, from the network input to the output. The best results in each column are in bold.}
\centering
\begin{tabular}{ccccc}
\hline
\textbf{Frozen Blocks} & \multicolumn{2}{c}{\textbf{No Pooling}} & \multicolumn{2}{c}{\textbf{Average Pooling}} \\
\hline
None    & $89.19\%$ & $\pm 6.46\%$              & $86.93\%$ & $\pm 6.42\%$  \\
1       & $88.56\%$ & $\pm 5.84\%$              & $88.95\%$ & $\pm 6.84\%$  \\
2       & $88.96\%$ & $\pm 5.99\%$              & $\bf{89.23\%}$ & $\bf{\pm 7.52\%}$  \\
3       & $\bf{89.40\%}$ & $\bf{\pm 6.50\%}$    & $88.42\%$ & $\pm 6.84\%$  \\
4       & $87.10\%$ & $\pm 5.84\%$              & $87.65\%$ & $\pm 7.39\%$  \\
All     & $86.36\%$ & $\pm 7.09\%$              & $74.78\%$ & $\pm 7.00\%$  \\
\hline
\end{tabular}
\label{tab:TLVGG16}
\end{table}

Xception and MobileNet have $13$ and $14$ layer blocks, respectively. Table \ref{tab:TLXceptionMobileNet} shows the classification accuracy with different amounts of frozen blocks and global average pooling. All the results were obtained using K-Fold Cross Validation with $k=10$; the whole process is repeated $2$ times, so each value on the table is the average of $20$ executions. Both models only learn when all layers are fine-tuned.

\begin{table}
\caption{Classification accuracy obtained using transfer learning with Xception and MobileNet and different amounts of frozen blocks of layers, from the network input to the output. The best results in each column are in bold.}
\centering
\begin{tabular}{ccccc}
\hline
\textbf{Frozen Blocks} & \multicolumn{2}{c}{\textbf{Xception}} & \multicolumn{2}{c}{\textbf{MobileNet}} \\
\hline
None    & $\bf{91.68\%}$ & $\bf{\pm 3.58\%}$  & $\bf{88.89\%}$ & $\bf{\pm 3.36\%}$  \\
1       & $48.26\%$ & $\pm 8.27\%$  & $51.17\%$ & $\pm 7.48\%$  \\
2       & $49.55\%$ & $\pm 8.82\%$  & $52.03\%$ & $\pm 9.70\%$  \\
3       & $48.95\%$ & $\pm 7.91\%$  & $50.74\%$ & $\pm 10.39\%$ \\
4       & $48.80\%$ & $\pm 7.35\%$  & $49.85\%$ & $\pm 10.05\%$ \\
5       & $51.18\%$ & $\pm 11.15\%$ & $45.13\%$ & $\pm 10.43\%$ \\
\hline
\end{tabular}
\label{tab:TLXceptionMobileNet}
\end{table}

\section{Semi-Supervised Learning with Particle Competition and Cooperation}
\label{sec:PCC}

CNNs inference in smartphones is feasible as long as the weights are set \cite{Simonyan2015}. However, the training process to set these weights is computationally intensive and requires more powerful hardware. Therefore, incorporating new knowledge, acquired from user feedback, through retraining is not feasible.

To address this issue, we propose an alternative approach by combining CNNs with PCC. In this scenario, VGG16 and VGG19 architectures are employed as feature extractors. They are used without their classification layer, with weights pre-trained on the ImageNet dataset \cite{ILSVRC15}. Those architectures were chosen among the others because they achieved much higher accuracy than the others on the proposed dataset in the scenario where no fine-tuning of the weights is used, as shown in Table \ref{tab:TLallCNN}. The last convolutional layer of both VGG16 and VGG19 architectures output a $7 \times 7 \times 512$ matrix. Therefore, with no pooling, there is a total of $25,088$ features. Alternatively, global average pooling or global max pooling may be used to lower it to $512$ features.

Principal Component Analysis (PCA) \cite{Jolliffe2002} is used to reduce the dimensionality. In preliminary experiments, we noticed that no more than $20$ of the principal components are needed to achieve the best results in most scenarios. The principal components are used to build an unweighted and undirect graph, in which each image is a node and edges connect each image to their $k$-nearest neighbors, according to the Euclidean distance between the principal components being used.

The graph is then fed to PCC with a few annotated images and it performs the classification of the remaining. PCC computational complexity is only $O(n)$ where $n$ is the amount of images \cite{Breve2012}. Therefore, it is suitable for fast execution on smartphones. Though the original PCC model is transductive, it was already extended to perform inductive learning and to work with data streams \cite{Breve2012IJCNN}, as the problem requires.

\begin{figure}
  \centering
  \includegraphics[width=\columnwidth]{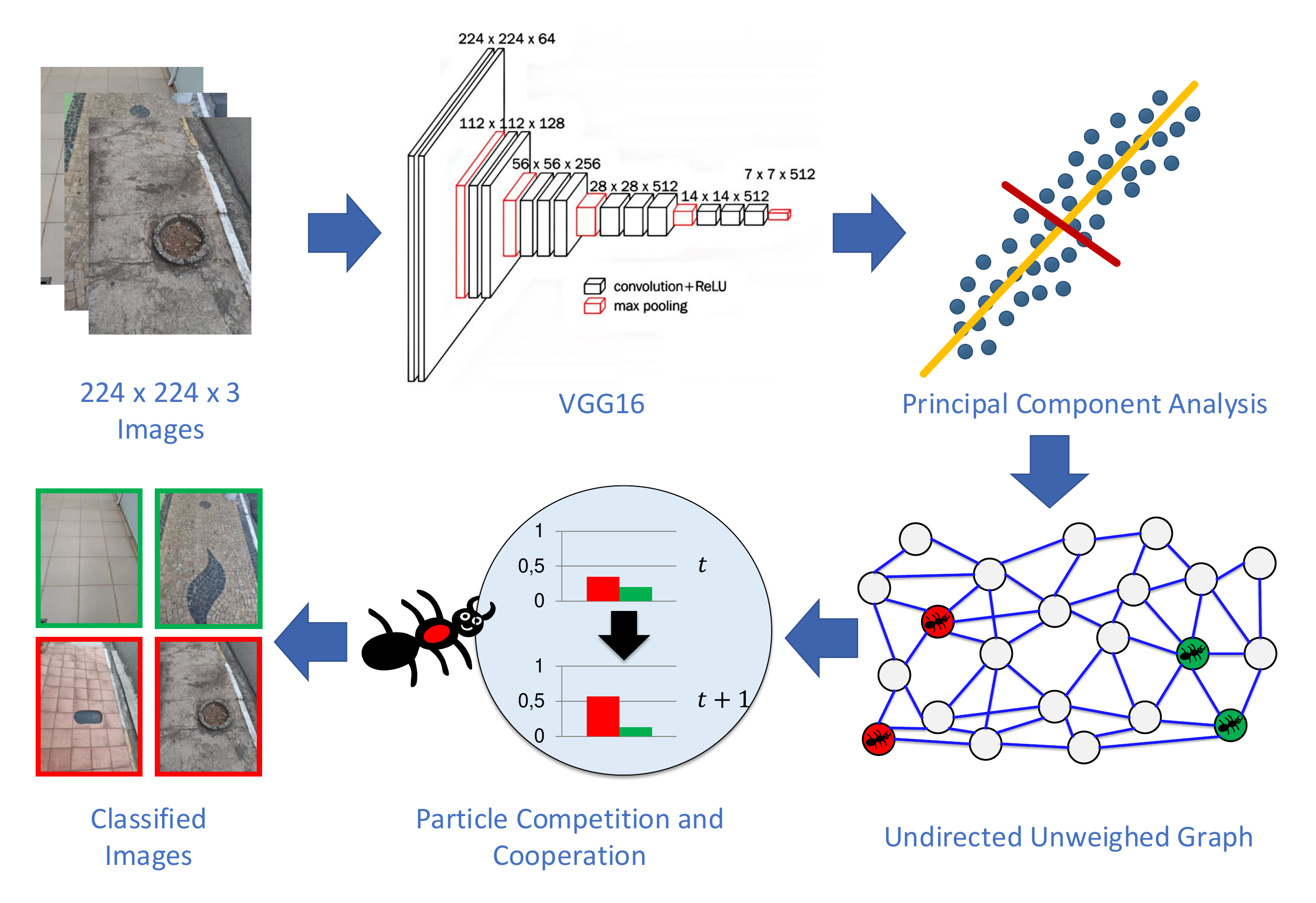}
  \caption{The proposed framework using Particle Competition and Cooperation for semi-supervised classification with VGG16 as feature extractor.}
  \label{fig:VGG-PCC-Diagram}
\end{figure}

Fig.~\ref{fig:VGG-PCC-Diagram} illustrates the classification framework using VGG16 as feature extractor and PCC as the semi-supervised classifier.

\begin{figure}
\setlength\tabcolsep{0pt}
\centering
\begin{subfigure}[b]{.49\columnwidth}
    \centering
    \includegraphics[width=\linewidth]{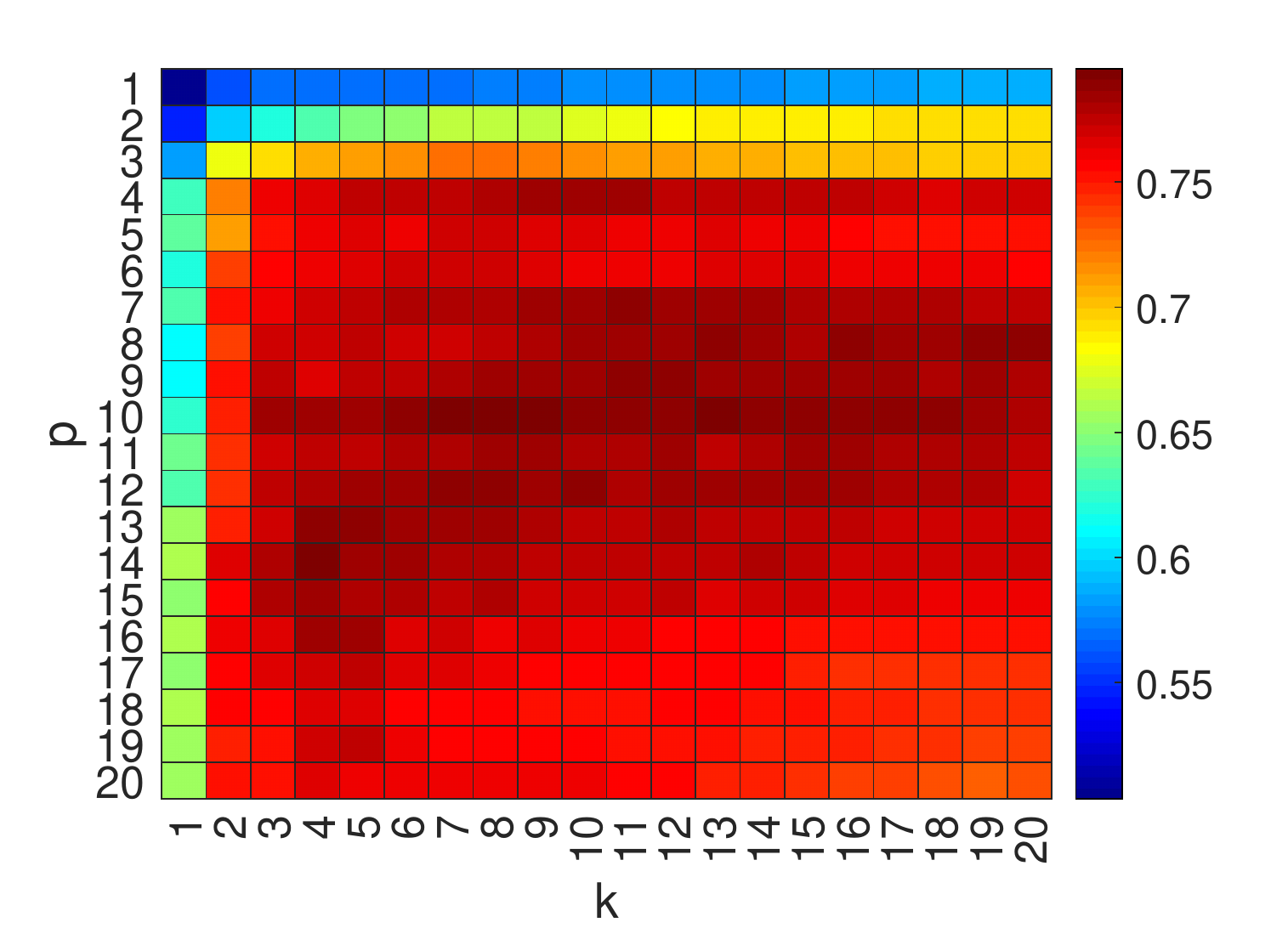}
    \caption{VGG16 - $10\%$}
\end{subfigure}
\begin{subfigure}[b]{.49\columnwidth}
    \centering
    \includegraphics[width=\linewidth]{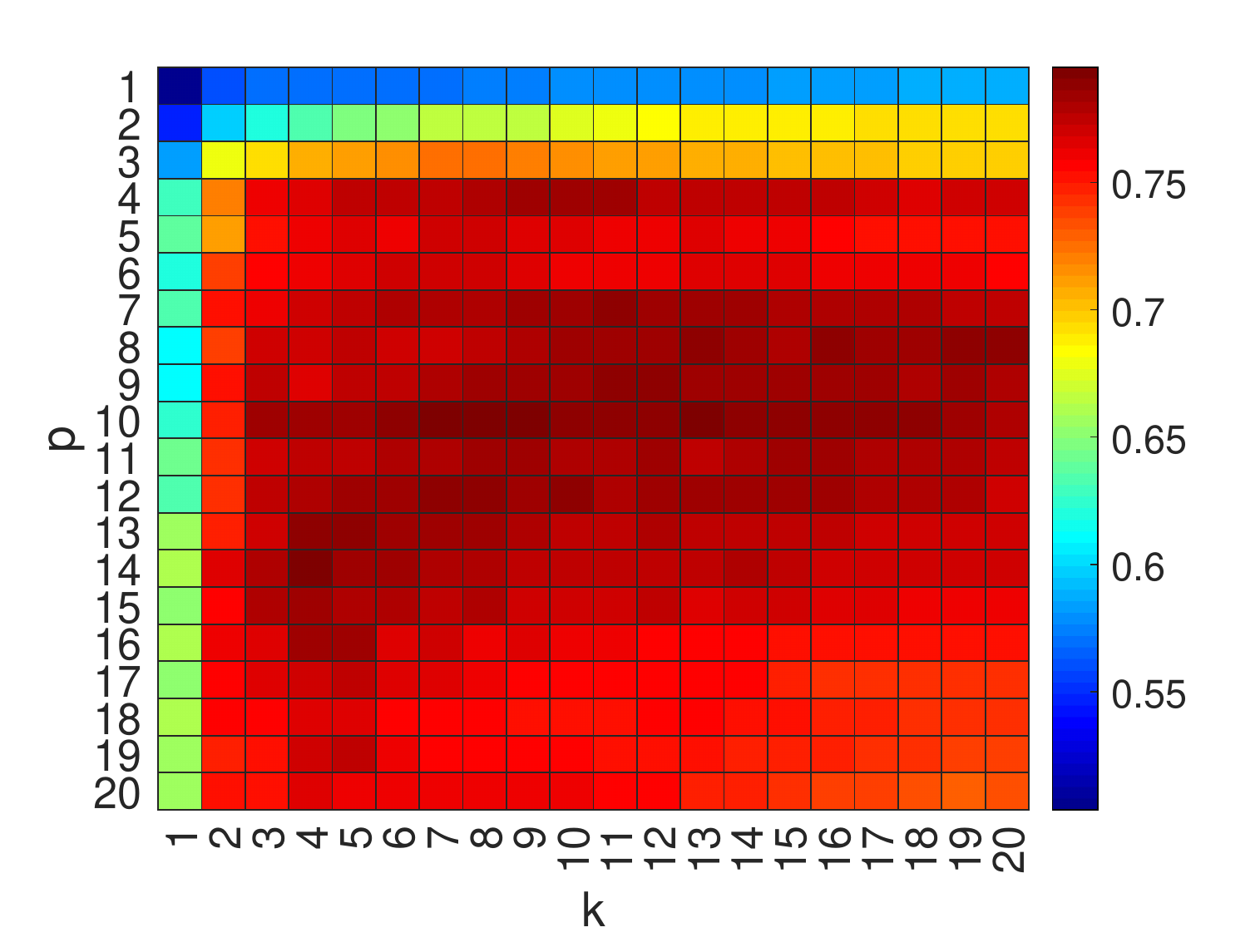}
    \caption{VGG16 - $20\%$}
\end{subfigure}

\begin{subfigure}[b]{.49\columnwidth}
    \centering
    \includegraphics[width=\linewidth]{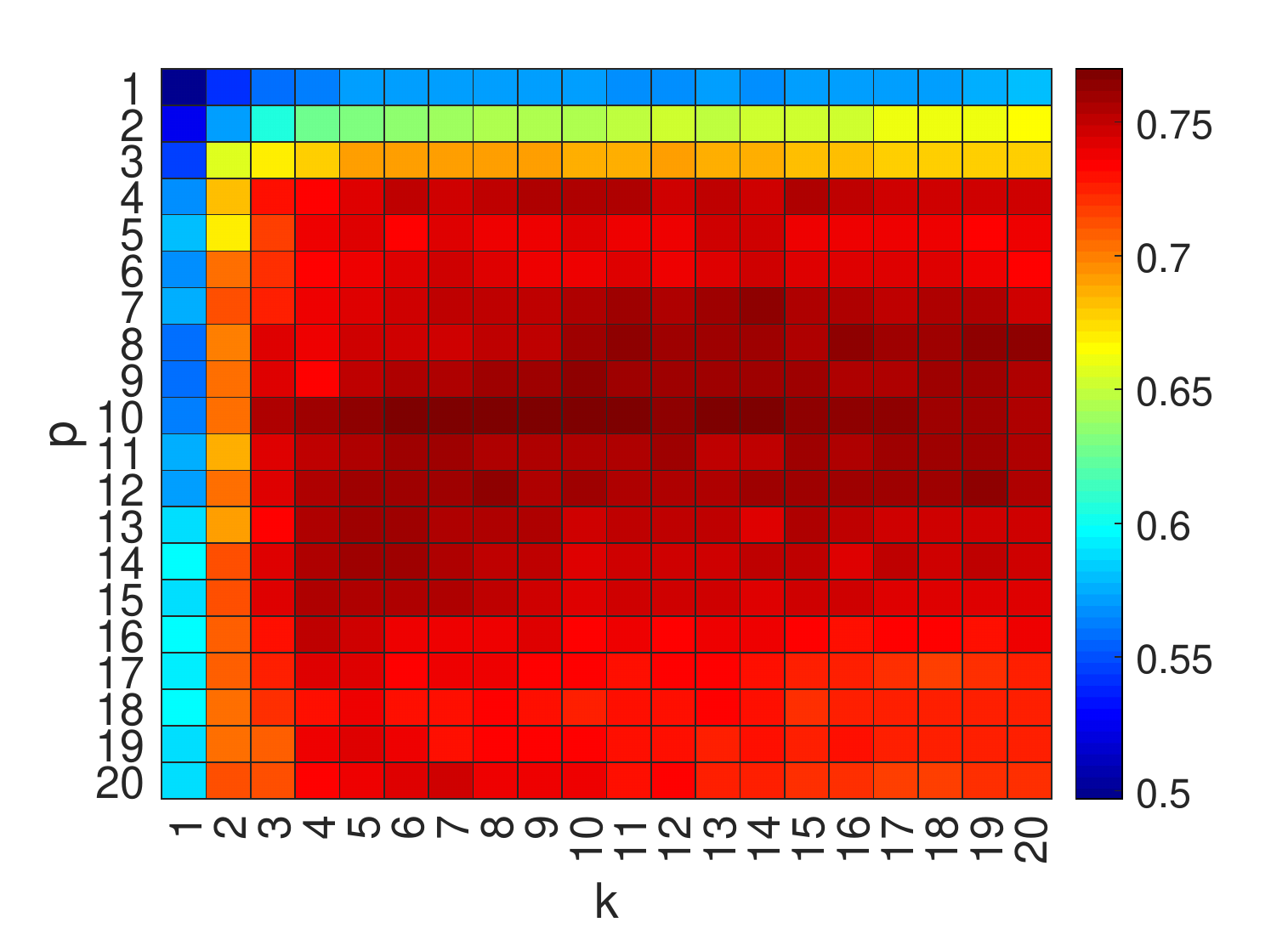}
    \caption{VGG19 - $10\%$}
\end{subfigure}
\begin{subfigure}[b]{.49\columnwidth}
    \centering
    \includegraphics[width=\linewidth]{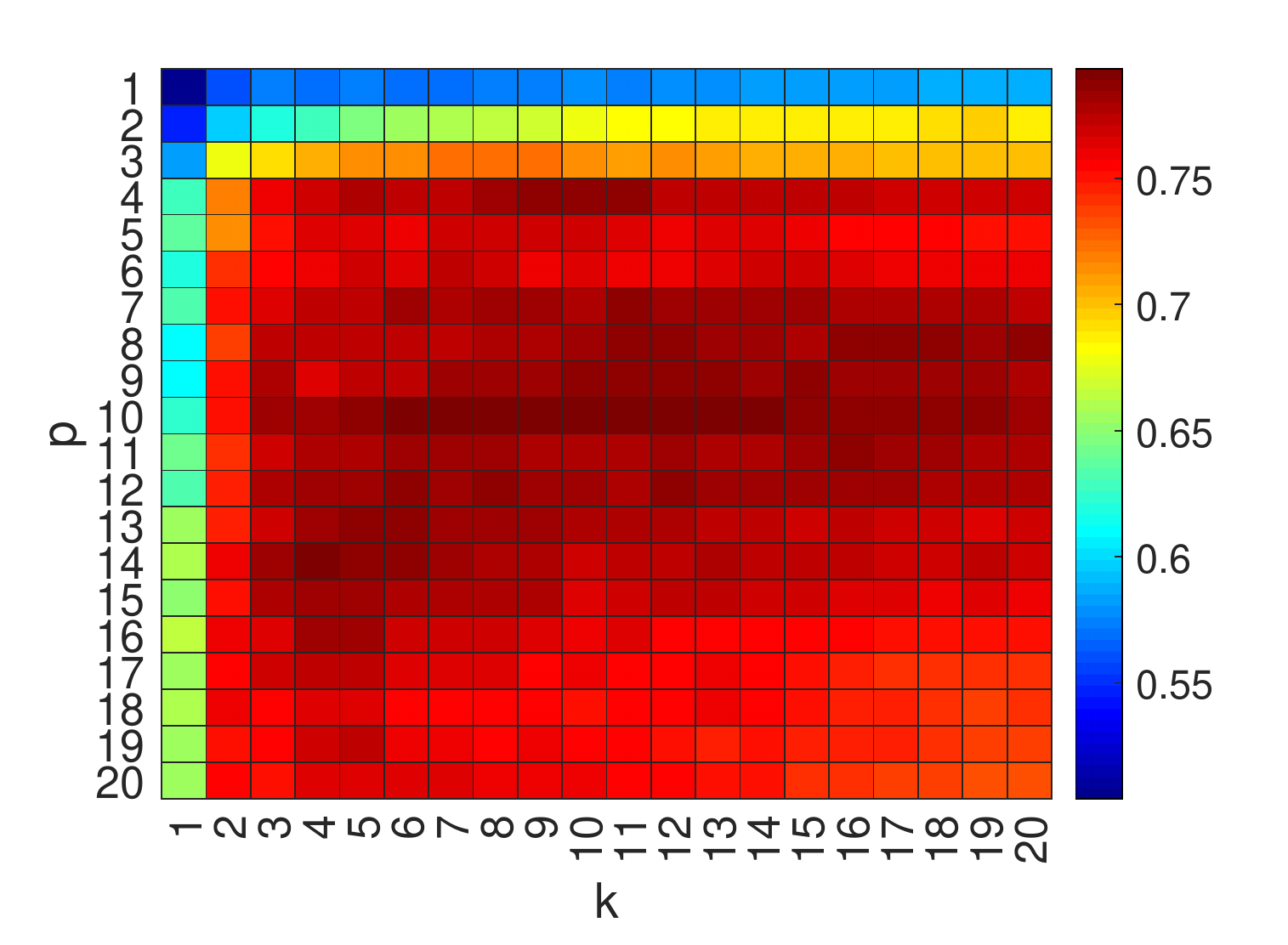}
    \caption{VGG19 - $20\%$}
\end{subfigure}

\begin{subfigure}[b]{.49\columnwidth}
    \centering
    \includegraphics[width=\linewidth]{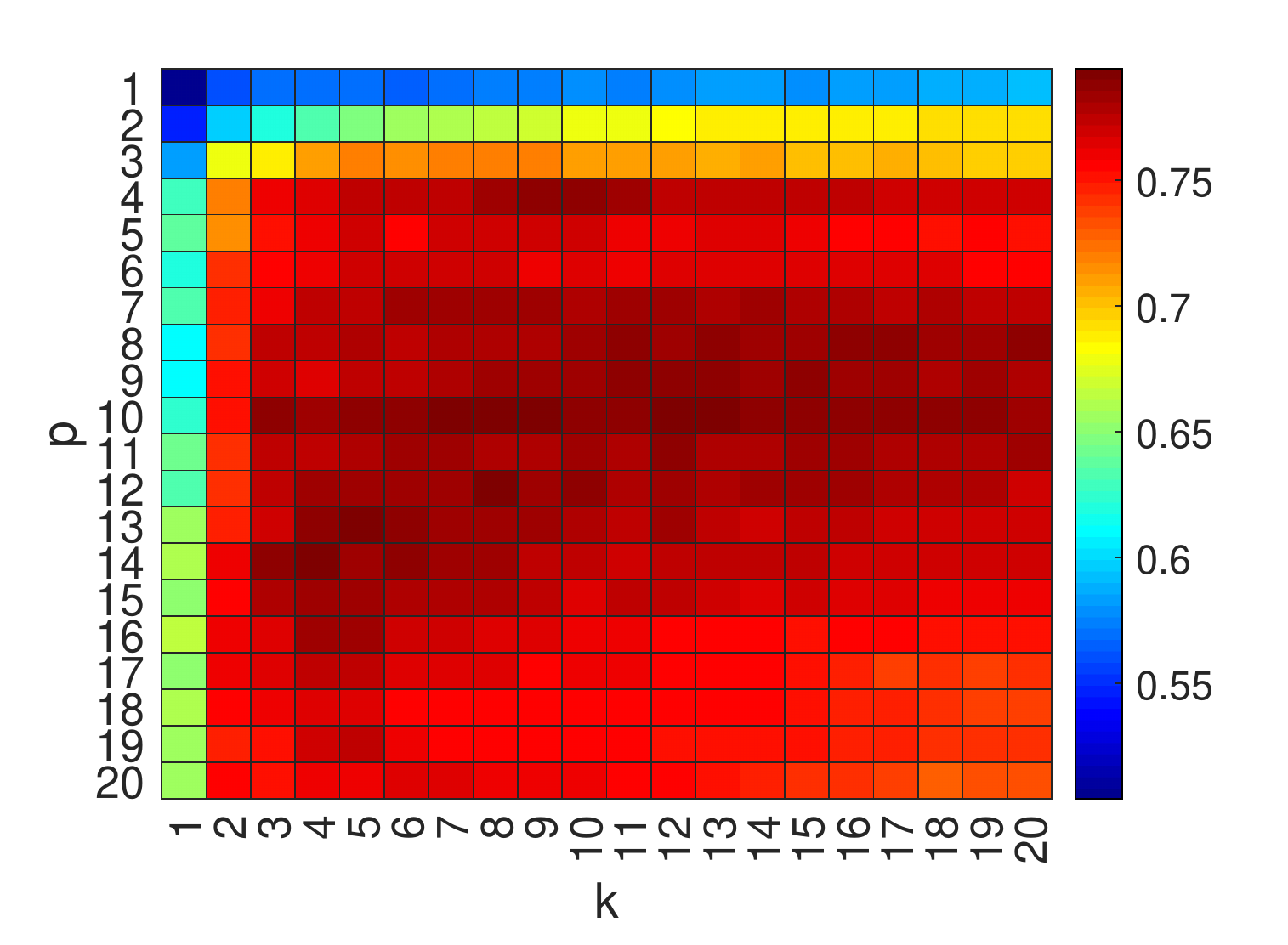}
    \caption{VGG16+VGG19 - $10\%$}
\end{subfigure}
\begin{subfigure}[b]{.49\columnwidth}
    \centering
    \includegraphics[width=\linewidth]{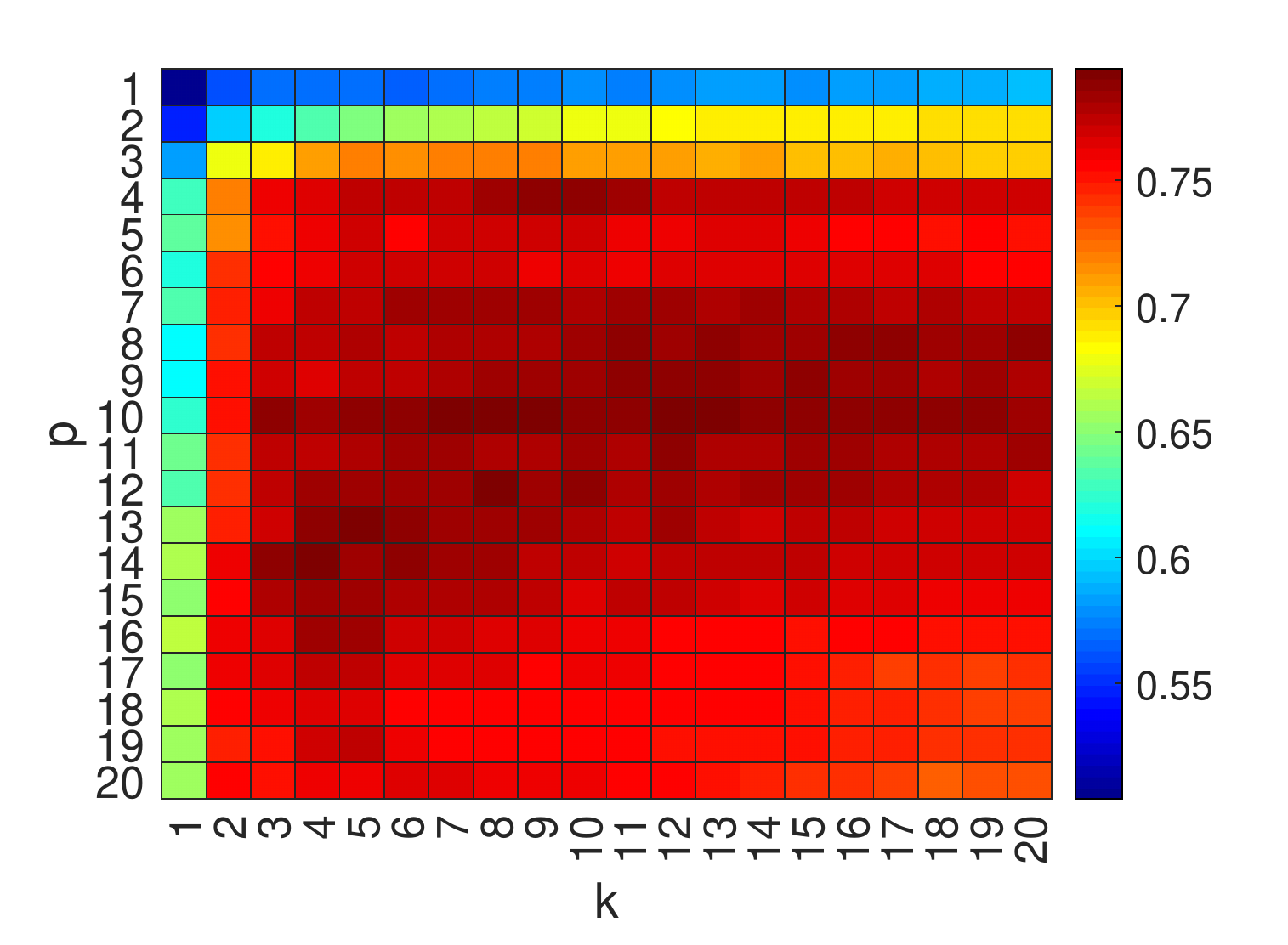}
    \caption{VGG16+VGG19 - $20\%$}
\end{subfigure}
\caption{PCC framework accuracy varying the amount of $k$-nearest neighbors and $p$ principal components: (a) VGG16 and $10\%$ labeled nodes; (b) VGG16 and $20\%$ labeled nodes; (c) VGG19 and $10\%$ labeled nodes; (d) VGG19 and $20\%$ labeled nodes; (e) VGG16+VGG19 and $10\%$ labeled nodes; (f) VGG16+VGG19 and $20\%$ labeled nodes.}
\label{fig:PCCHeatmaps}
\end{figure}

Fig.~\ref{fig:PCCHeatmaps} shows the accuracy of the PCC framework on the proposed dataset with VGG16 and VGG19 used as feature extractors. They were also tested together, by concatenating their output, generating $50,176$ features. In these simulations, we used the output of their last convolutional layer, with no pooling. $10\%$ and $20\%$ of the images are presented with their respective labels; so the remaining are labeled by PCC. The optimal amount of $k$-nearest neighbors and $p$ principal components to build the graph are found by using grid-search in the interval $k,p = \{1, \ldots, 20\}$. The PCA MATLAB implementation is used with its default parameters. PCC is implemented in MATLAB and all its parameters are left at their default values\footnote{PCC MATLAB implementation is available at \url{https://github.com/fbreve/Particle-Competition-and-Cooperation}}. Table \ref{tab:PCC} shows the best combination of $k$ and $p$ for each scenario and the accuracy they achieved. All combinations are repeated $100$ times with different randomly chosen labeled nodes, so the values on the Fig.~\ref{fig:PCCHeatmaps} and Table \ref{tab:PCC} are the average of $100$ executions.

\begin{table}
\caption{Classification accuracy achieved by the proposed PCC framework with VGG16 and VGG19 as feature extractors, with the best combination of $p$ principal components and $k$-nearest neighbors to build the graph. No pooling is applied to the last convolutional layers output.}
\centering
\begin{tabular}{ccccrl}
\hline
{\bf Labeled} & {\bf Architecture} & {\bf $p$} & {\bf $k$} & \multicolumn{2}{c}{\bf Accuracy} \\
\hline
$10\%$ & VGG16          & $10$  & $7$   & $77.01\%$ & $ \pm 3.55\%$ \\
$10\%$ & VGG19          & $10$  & $8$   & $76.99\%$ & $ \pm 3.60\%$ \\
$10\%$ & VGG16+VGG19    & $10$  & $8$   & $76.99\%$ & $ \pm 3.68\%$ \\
\hline
$20\%$ & VGG16          & $10$  & $7$   & $79.53\%$ & $ \pm 2.40\%$ \\
$20\%$ & VGG19          & $10$  & $8$   & $79.35\%$ & $ \pm 2.65\%$ \\
$20\%$ & VGG16+VGG19    & $14$  & $4$   & $79.43\%$ & $ \pm 2.65\%$ \\
\hline
\end{tabular}
\label{tab:PCC}
\end{table}

By analyzing Fig.~\ref{fig:PCCHeatmaps} and Table \ref{tab:PCC}, we notice that $10$ principal components is the optimal amount in five of the tested scenarios, which is considerably lesser than the original thousands of features. We then repeated the tests with $20\%$ labeled items but now using global average pooling and global max pooling, to lower the initial amount of features to $512$, before applying PCA. However, in these scenarios, the accuracy lowered, as shown in Table \ref{tab:PCCPooling}, indicating that useful discriminative information is lost when pooling is used.

\begin{table}
\caption{Classification accuracy achieved by the proposed PCC framework with VGG16 and VGG19 as feature extractors, with global pooling applied to their last convolutional layer output. The best combination of $p$ principal components and $k$-nearest neighbors are used. $20\%$ of the data items are randomly labeled. All values are the average of $100$ executions.}
\centering
\begin{tabular}{ccccrl}
\hline
{\bf Global Pooling} & {\bf Architecture} & {\bf $p$} & {\bf $k$} & \multicolumn{2}{c}{\bf Accuracy} \\
\hline
Average & VGG16         & $7$  & $6$ & $72.51\%$ & $ \pm 3.04\%$ \\
Average & VGG19         & $15$ & $3$ & $71.52\%$ & $ \pm 3.28\%$ \\
Average & VGG16+VGG19   & $10$ & $6$ & $73.43\%$ & $ \pm 3.10\%$ \\
\hline
Max     & VGG16         & $7$   & $7$ & $74.30\%$ & $ \pm 2.80\%$ \\
Max     & VGG19         & $20$  & $8$ & $72.28\%$ & $ \pm 3.87\%$ \\
Max     & VGG16+VGG19   & $20$  & $4$ & $73.19\%$ & $ \pm 3.35\%$ \\
\hline
\end{tabular}
\label{tab:PCCPooling}
\end{table}

The results obtained with SSL are impressive considering that the CNNs were not fine-tuned and that only $10\%$ to $20\%$ of the images on the dataset were presented with the corresponding labels. In real-world scenarios, we could certainly use some images to fine-tune the networks before using them as feature extractors, increasing the extraction quality and, consequently, the classification accuracy.

\section{Conclusions}
\label{sec:Conclusions}

In this paper, we propose methods to help in identifying obstacles in the path of visually impaired people. After properly trained, these methods have low computational costs in the inference step, executing in milliseconds in current smartphones \cite{Ignatov2019}. Therefore, it can be implemented without relying on other equipment or remote servers. We also propose a dataset to help in the training of these methods. Our first identification approach uses CNNs and transfer learning. We compared many consolidated CNN architectures pre-trained on large datasets and fine-tuned them to the proposed task. The second approach uses the pre-trained CNN architectures as feature extractors and semi-supervised learning for classification, using the particle competition and cooperation method.

Computer simulations showed promising results with some of the CNN architectures, while the SSL also achieved relatively high accuracy, considering that it is using only up to $20\%$ of the dataset for training and no fine-tuning in CNN networks. As future work, we intend to acquire more images to the proposed dataset and search for other approaches and tweaks in the current framework to further improve the classification accuracy. Then, a smartphone prototype application may be implemented to test some real-world scenarios.


\bibliographystyle{IEEEtran}
\bibliography{IEEEabrv,wcci2020}

\begin{thebibliography}{10}
\providecommand{\url}[1]{#1}
\csname url@samestyle\endcsname
\providecommand{\newblock}{\relax}
\providecommand{\bibinfo}[2]{#2}
\providecommand{\BIBentrySTDinterwordspacing}{\spaceskip=0pt\relax}
\providecommand{\BIBentryALTinterwordstretchfactor}{4}
\providecommand{\BIBentryALTinterwordspacing}{\spaceskip=\fontdimen2\font plus
\BIBentryALTinterwordstretchfactor\fontdimen3\font minus
  \fontdimen4\font\relax}
\providecommand{\BIBforeignlanguage}[2]{{%
\expandafter\ifx\csname l@#1\endcsname\relax
\typeout{** WARNING: IEEEtran.bst: No hyphenation pattern has been}%
\typeout{** loaded for the language `#1'. Using the pattern for}%
\typeout{** the default language instead.}%
\else
\language=\csname l@#1\endcsname
\fi
#2}}
\providecommand{\BIBdecl}{\relax}
\BIBdecl

\bibitem{WHO2019}
\BIBentryALTinterwordspacing
{World Health Organization}, ``Vision impairment and blindness,'' Oct 2019,
  accessed: 2019-01-14. [Online]. Available:
  \url{https://www.who.int/news-room/fact-sheets/detail/blindness-and-visual-impairment}
\BIBentrySTDinterwordspacing

\bibitem{Bourne2017}
R.~R. Bourne, S.~R. Flaxman, T.~Braithwaite, M.~V. Cicinelli, A.~Das, J.~B.
  Jonas, J.~Keeffe, J.~H. Kempen, J.~Leasher, H.~Limburg \emph{et~al.},
  ``Magnitude, temporal trends, and projections of the global prevalence of
  blindness and distance and near vision impairment: a systematic review and
  meta-analysis,'' \emph{The Lancet Global Health}, vol.~5, no.~9, pp.
  e888--e897, 2017.

\bibitem{Lakde2015}
C.~K. Lakde and P.~S. Prasad, ``Review paper on navigation system for visually
  impaired people,'' \emph{International Journal of Advanced Research in
  Computer and Communication Engineering}, vol.~4, no.~1, 2015.

\bibitem{Jiang2019}
B.~{Jiang}, J.~{Yang}, Z.~{Lv}, and H.~{Song}, ``Wearable vision assistance
  system based on binocular sensors for visually impaired users,'' \emph{IEEE
  Internet of Things Journal}, vol.~6, no.~2, pp. 1375--1383, April 2019.

\bibitem{Hoang2017}
\BIBentryALTinterwordspacing
V.-N. Hoang, T.-H. Nguyen, T.-L. Le, T.-H. Tran, T.-P. Vuong, and N.~Vuillerme,
  ``Obstacle detection and warning system for visually impaired people based on
  electrode matrix and mobile kinect,'' \emph{Vietnam Journal of Computer
  Science}, vol.~4, no.~2, pp. 71--83, May 2017. [Online]. Available:
  \url{https://doi.org/10.1007/s40595-016-0075-z}
\BIBentrySTDinterwordspacing

\bibitem{Tapu2017}
R.~Tapu, B.~Mocanu, and T.~Zaharia, ``Deep-see: Joint object detection,
  tracking and recognition with application to visually impaired navigational
  assistance,'' \emph{Sensors}, vol.~17, no.~11, p. 2473, 2017.

\bibitem{Saffoury2016}
R.~Saffoury, P.~Blank, J.~Sessner, B.~H. Groh, C.~F. Martindale, E.~Dorschky,
  J.~Franke, and B.~M. Eskofier, ``Blind path obstacle detector using
  smartphone camera and line laser emitter,'' in \emph{2016 1st International
  Conference on Technology and Innovation in Sports, Health and Wellbeing
  (TISHW)}.\hskip 1em plus 0.5em minus 0.4em\relax IEEE, 2016, pp. 1--7.

\bibitem{Poggi2016}
M.~Poggi and S.~Mattoccia, ``A wearable mobility aid for the visually impaired
  based on embedded 3d vision and deep learning,'' in \emph{2016 IEEE Symposium
  on Computers and Communication (ISCC)}.\hskip 1em plus 0.5em minus
  0.4em\relax IEEE, 2016, pp. 208--213.

\bibitem{Poggi2015}
M.~Poggi, L.~Nanni, and S.~Mattoccia, ``Crosswalk recognition through
  point-cloud processing and deep-learning suited to a wearable mobility aid
  for the visually impaired,'' in \emph{International Conference on Image
  Analysis and Processing}.\hskip 1em plus 0.5em minus 0.4em\relax Springer,
  2015, pp. 282--289.

\bibitem{Rizzo2017}
J.-R. Rizzo, Y.~Pan, T.~Hudson, E.~K. Wong, and Y.~Fang, ``Sensor fusion for
  ecologically valid obstacle identification: Building a comprehensive
  assistive technology platform for the visually impaired,'' in \emph{2017 7th
  International Conference on Modeling, Simulation, and Applied Optimization
  (ICMSAO)}.\hskip 1em plus 0.5em minus 0.4em\relax IEEE, 2017, pp. 1--5.

\bibitem{Lin2017}
B.-S. Lin, C.-C. Lee, and P.-Y. Chiang, ``Simple smartphone-based guiding
  system for visually impaired people,'' \emph{Sensors}, vol.~17, no.~6, p.
  1371, 2017.

\bibitem{Lecun2015}
Y.~LeCun, Y.~Bengio, and G.~Hinton, ``Deep learning,'' \emph{nature}, vol. 521,
  no. 7553, p. 436, 2015.

\bibitem{Goodfellow2016}
I.~Goodfellow, Y.~Bengio, and A.~Courville, \emph{Deep learning}.\hskip 1em
  plus 0.5em minus 0.4em\relax MIT press, 2016.

\bibitem{Schmidhuber2015}
J.~Schmidhuber, ``Deep learning in neural networks: An overview,'' \emph{Neural
  networks}, vol.~61, pp. 85--117, 2015.

\bibitem{Krizhevsky2012}
\BIBentryALTinterwordspacing
A.~Krizhevsky, I.~Sutskever, and G.~E. Hinton, ``Imagenet classification with
  deep convolutional neural networks,'' in \emph{Advances in Neural Information
  Processing Systems 25}, F.~Pereira, C.~J.~C. Burges, L.~Bottou, and K.~Q.
  Weinberger, Eds.\hskip 1em plus 0.5em minus 0.4em\relax Curran Associates,
  Inc., 2012, pp. 1097--1105. [Online]. Available:
  \url{http://papers.nips.cc/paper/4824-imagenet-classification-with-deep-convolutional-neural-networks.pdf}
\BIBentrySTDinterwordspacing

\bibitem{Cai2016}
Z.~Cai, Q.~Fan, R.~S. Feris, and N.~Vasconcelos, ``A unified multi-scale deep
  convolutional neural network for fast object detection,'' in \emph{european
  conference on computer vision}.\hskip 1em plus 0.5em minus 0.4em\relax
  Springer, 2016, pp. 354--370.

\bibitem{Howard2017}
A.~G. Howard, M.~Zhu, B.~Chen, D.~Kalenichenko, W.~Wang, T.~Weyand,
  M.~Andreetto, and H.~Adam, ``Mobilenets: Efficient convolutional neural
  networks for mobile vision applications,'' \emph{arXiv preprint
  arXiv:1704.04861}, 2017.

\bibitem{Oquab2014}
M.~Oquab, L.~Bottou, I.~Laptev, and J.~Sivic, ``Learning and transferring
  mid-level image representations using convolutional neural networks,'' in
  \emph{The IEEE Conference on Computer Vision and Pattern Recognition (CVPR)},
  June 2014.

\bibitem{Shin2016}
H.~{Shin}, H.~R. {Roth}, M.~{Gao}, L.~{Lu}, Z.~{Xu}, I.~{Nogues}, J.~{Yao},
  D.~{Mollura}, and R.~M. {Summers}, ``Deep convolutional neural networks for
  computer-aided detection: Cnn architectures, dataset characteristics and
  transfer learning,'' \emph{IEEE Transactions on Medical Imaging}, vol.~35,
  no.~5, pp. 1285--1298, May 2016.

\bibitem{Huynh2016}
\BIBentryALTinterwordspacing
B.~Q. Huynh, H.~Li, and M.~L. Giger, ``{Digital mammographic tumor
  classification using transfer learning from deep convolutional neural
  networks},'' \emph{Journal of Medical Imaging}, vol.~3, no.~3, pp. 1 -- 5,
  2016. [Online]. Available: \url{https://doi.org/10.1117/1.JMI.3.3.034501}
\BIBentrySTDinterwordspacing

\bibitem{Gopalakrishnan2017}
\BIBentryALTinterwordspacing
K.~Gopalakrishnan, S.~K. Khaitan, A.~Choudhary, and A.~Agrawal, ``Deep
  convolutional neural networks with transfer learning for computer
  vision-based data-driven pavement distress detection,'' \emph{Construction
  and Building Materials}, vol. 157, pp. 322 -- 330, 2017. [Online]. Available:
  \url{http://www.sciencedirect.com/science/article/pii/S0950061817319335}
\BIBentrySTDinterwordspacing

\bibitem{Simonyan2015}
K.~Simonyan and A.~Zisserman, ``Very deep convolutional networks for
  large-scale image recognition.''\hskip 1em plus 0.5em minus 0.4em\relax
  Computational and Biological Learning Society, 2015, pp. 1--14.

\bibitem{Szegedy2016}
C.~Szegedy, V.~Vanhoucke, S.~Ioffe, J.~Shlens, and Z.~Wojna, ``Rethinking the
  inception architecture for computer vision,'' in \emph{The IEEE Conference on
  Computer Vision and Pattern Recognition (CVPR)}, June 2016, pp. 2818--2826.

\bibitem{Ignatov2019}
A.~Ignatov and R.~Timofte, ``Ai benchmark: All about deep learning on
  smartphones in 2019,'' in \emph{IEEE International Conference on Computer
  Vision (ICCV) Workshops}, 2019.

\bibitem{ILSVRC15}
O.~Russakovsky, J.~Deng, H.~Su, J.~Krause, S.~Satheesh, S.~Ma, Z.~Huang,
  A.~Karpathy, A.~Khosla, M.~Bernstein, A.~C. Berg, and L.~Fei-Fei, ``{ImageNet
  Large Scale Visual Recognition Challenge},'' \emph{International Journal of
  Computer Vision (IJCV)}, vol. 115, no.~3, pp. 211--252, 2015.

\bibitem{Chollet2017}
F.~{Chollet}, ``Xception: Deep learning with depthwise separable
  convolutions,'' in \emph{2017 IEEE Conference on Computer Vision and Pattern
  Recognition (CVPR)}, July 2017, pp. 1800--1807.

\bibitem{Breve2012}
F.~{Breve}, L.~{Zhao}, M.~{Quiles}, W.~{Pedrycz}, and J.~{Liu}, ``Particle
  competition and cooperation in networks for semi-supervised learning,''
  \emph{IEEE Transactions on Knowledge and Data Engineering}, vol.~24, no.~9,
  pp. 1686--1698, Sep. 2012.

\bibitem{Jolliffe2002}
I.~Jolliffe and Springer-Verlag, \emph{Principal Component Analysis}, ser.
  Springer Series in Statistics.\hskip 1em plus 0.5em minus 0.4em\relax
  Springer, 2002.

\bibitem{Kumar2015}
R.~{Kumar} and S.~{Meher}, ``A novel method for visually impaired using object
  recognition,'' in \emph{2015 International Conference on Communications and
  Signal Processing (ICCSP)}, April 2015, pp. 0772--0776.

\bibitem{Tapu2013}
R.~Tapu, B.~Mocanu, A.~Bursuc, and T.~Zaharia, ``A smartphone-based obstacle
  detection and classification system for assisting visually impaired people,''
  in \emph{The IEEE International Conference on Computer Vision (ICCV)
  Workshops}, June 2013.

\bibitem{Islam2018}
M.~M. {Islam} and M.~S. {Sadi}, ``Path hole detection to assist the visually
  impaired people in navigation,'' in \emph{2018 4th International Conference
  on Electrical Engineering and Information Communication Technology
  (iCEEiCT)}, Sep. 2018, pp. 268--273.

\bibitem{Everingham2010}
\BIBentryALTinterwordspacing
M.~Everingham, L.~Van~Gool, C.~K.~I. Williams, J.~Winn, and A.~Zisserman, ``The
  pascal visual object classes (voc) challenge,'' \emph{International Journal
  of Computer Vision}, vol.~88, no.~2, pp. 303--338, Jun 2010. [Online].
  Available: \url{https://doi.org/10.1007/s11263-009-0275-4}
\BIBentrySTDinterwordspacing

\bibitem{Saleh2017}
K.~{Saleh}, R.~A. {Zeineldin}, M.~{Hossny}, S.~{Nahavandi}, and N.~A.
  {El-Fishawy}, ``Navigational path detection for the visually impaired using
  fully convolutional networks,'' in \emph{2017 IEEE International Conference
  on Systems, Man, and Cybernetics (SMC)}, Oct 2017, pp. 1399--1404.

\bibitem{Monteiro2017}
J.~{Monteiro}, J.~P. {Aires}, R.~{Granada}, R.~C. {Barros}, and F.~{Meneguzzi},
  ``Virtual guide dog: An application to support visually-impaired people
  through deep convolutional neural networks,'' in \emph{2017 International
  Joint Conference on Neural Networks (IJCNN)}, May 2017, pp. 2267--2274.

\bibitem{Szegedy2015}
C.~Szegedy, W.~Liu, Y.~Jia, P.~Sermanet, S.~Reed, D.~Anguelov, D.~Erhan,
  V.~Vanhoucke, and A.~Rabinovich, ``Going deeper with convolutions,'' in
  \emph{The IEEE Conference on Computer Vision and Pattern Recognition (CVPR)},
  June 2015.

\bibitem{Breve2013SoftComputing}
\BIBentryALTinterwordspacing
F.~Breve and L.~Zhao, ``Fuzzy community structure detection by particle
  competition and cooperation,'' \emph{Soft Computing}, vol.~17, no.~4, pp.
  659--673, Apr 2013. [Online]. Available:
  \url{https://doi.org/10.1007/s00500-012-0924-3}
\BIBentrySTDinterwordspacing

\bibitem{Breve2015Neurocomputing}
\BIBentryALTinterwordspacing
F.~A. Breve, L.~Zhao, and M.~G. Quiles, ``Particle competition and cooperation
  for semi-supervised learning with label noise,'' \emph{Neurocomputing}, vol.
  160, pp. 63 -- 72, 2015. [Online]. Available:
  \url{http://www.sciencedirect.com/science/article/pii/S0925231215001277}
\BIBentrySTDinterwordspacing

\bibitem{Breve2012IJCNN}
F.~{Breve} and L.~{Zhao}, ``Particle competition and cooperation in networks
  for semi-supervised learning with concept drift,'' in \emph{The 2012
  International Joint Conference on Neural Networks (IJCNN)}, June 2012, pp.
  1--6.

\bibitem{Breve2013IJCNN}
F.~{Breve}, ``Active semi-supervised learning using particle competition and
  cooperation in networks,'' in \emph{The 2013 International Joint Conference
  on Neural Networks (IJCNN)}, Aug 2013, pp. 1--6.

\bibitem{Breve2015IJCNN}
F.~{Breve}, M.~G. {Quiles}, and L.~{Zhao}, ``Interactive image segmentation
  using particle competition and cooperation,'' in \emph{2015 International
  Joint Conference on Neural Networks (IJCNN)}, July 2015, pp. 1--8.

\bibitem{Kingma2014}
D.~P. Kingma and J.~Ba, ``Adam: A method for stochastic optimization,'' 2014.

\bibitem{Reddi2018}
\BIBentryALTinterwordspacing
S.~J. Reddi, S.~Kale, and S.~Kumar, ``On the convergence of adam and beyond,''
  in \emph{International Conference on Learning Representations}, 2018.
  [Online]. Available: \url{https://openreview.net/forum?id=ryQu7f-RZ}
\BIBentrySTDinterwordspacing

\bibitem{Tieleman2012}
T.~Tieleman and G.~Hinton, ``Lecture 6.5-rmsprop: Divide the gradient by a
  running average of its recent magnitude,'' \emph{COURSERA: Neural networks
  for machine learning}, vol.~4, no.~2, pp. 26--31, 2012.

\bibitem{He2016}
K.~He, X.~Zhang, S.~Ren, and J.~Sun, ``Deep residual learning for image
  recognition,'' in \emph{The IEEE Conference on Computer Vision and Pattern
  Recognition (CVPR)}, June 2016, pp. 770--778.

\bibitem{He2016ResNetV2}
------, ``Identity mappings in deep residual networks,'' in \emph{Computer
  Vision -- ECCV 2016}, B.~Leibe, J.~Matas, N.~Sebe, and M.~Welling, Eds.\hskip
  1em plus 0.5em minus 0.4em\relax Cham: Springer International Publishing,
  2016, pp. 630--645.

\bibitem{Szegedy2017}
C.~Szegedy, S.~Ioffe, V.~Vanhoucke, and A.~A. Alemi, ``Inception-v4,
  inception-resnet and the impact of residual connections on learning,'' in
  \emph{Thirty-first AAAI conference on artificial intelligence}, 2017.

\bibitem{Huang2017}
G.~Huang, Z.~Liu, L.~van~der Maaten, and K.~Q. Weinberger, ``Densely connected
  convolutional networks,'' in \emph{The IEEE Conference on Computer Vision and
  Pattern Recognition (CVPR)}, July 2017, pp. 4700--4708.

\bibitem{Zoph2018}
B.~Zoph, V.~Vasudevan, J.~Shlens, and Q.~V. Le, ``Learning transferable
  architectures for scalable image recognition,'' in \emph{The IEEE Conference
  on Computer Vision and Pattern Recognition (CVPR)}, June 2018, pp.
  8697--8710.

\bibitem{Sandler2018}
M.~Sandler, A.~Howard, M.~Zhu, A.~Zhmoginov, and L.-C. Chen, ``Mobilenetv2:
  Inverted residuals and linear bottlenecks,'' in \emph{The IEEE Conference on
  Computer Vision and Pattern Recognition (CVPR)}, June 2018, pp. 4510--4520.

\end{thebibliography}

\end{document}